\documentclass{article}
\usepackage{spconf,amsmath,graphicx}

\usepackage{amssymb}
\usepackage{booktabs}
\usepackage{url}

\def\etal{{\it et al.} }

\def\ie{{\it i.e.} }
\def\Cref{\ref}

\title{Texture- and Shape-based Adversarial Attacks\\for Overhead Image Vehicle Detection}
\name{
\begin{tabular}{@{}c@{}}
    Mikael Yeghiazaryan$^{1}$ \qquad 
    Sai Abhishek Siddhartha Namburu$^{1}$ \qquad 
    Emily Kim$^{1}$ \qquad
    Stanislav Panev$^{1}$\\ 
    Celso de Melo$^{2}$ \qquad 
    Fernando De la Torre$^{1}$ \qquad 
    Jessica K. Hodgins$^{1}$ 
\end{tabular}
}

\address{$^{1}$Carnegie Mellon University, USA \hspace{1cm}
         $^{2}$DEVCOM Army Research Lab}
\graphicspath{{Figures/}}

\usepackage[dvipsnames]{xcolor}

\usepackage{lipsum}

\usepackage{siunitx}
\usepackage{tabularx}

\usepackage{soul}
\usepackage{xcolor}
\usepackage{multirow}

\usepackage{caption}
\usepackage{subcaption}

\graphicspath{{Figures/}}

\usepackage{wrapfig}

\usepackage{subcaption}

\usepackage{mathtools}
\DeclarePairedDelimiter{\ceil}{\lceil}{\rceil}

\usepackage[pagebackref,breaklinks,colorlinks]{hyperref}
\usepackage[capitalize]{cleveref}
\crefname{section}{Sec.}{Secs.}
\Crefname{section}{Section}{Sections}
\Crefname{table}{Table}{Tables}
\crefname{table}{Tab.}{Tabs.}

\begin{document}
%\ninept
%
% \maketitle

\twocolumn[{%
\renewcommand\twocolumn[1][]{#1}%
\maketitle
\begin{center}
    \centering
    \captionsetup{type=figure}
    \includegraphics[width=\textwidth]{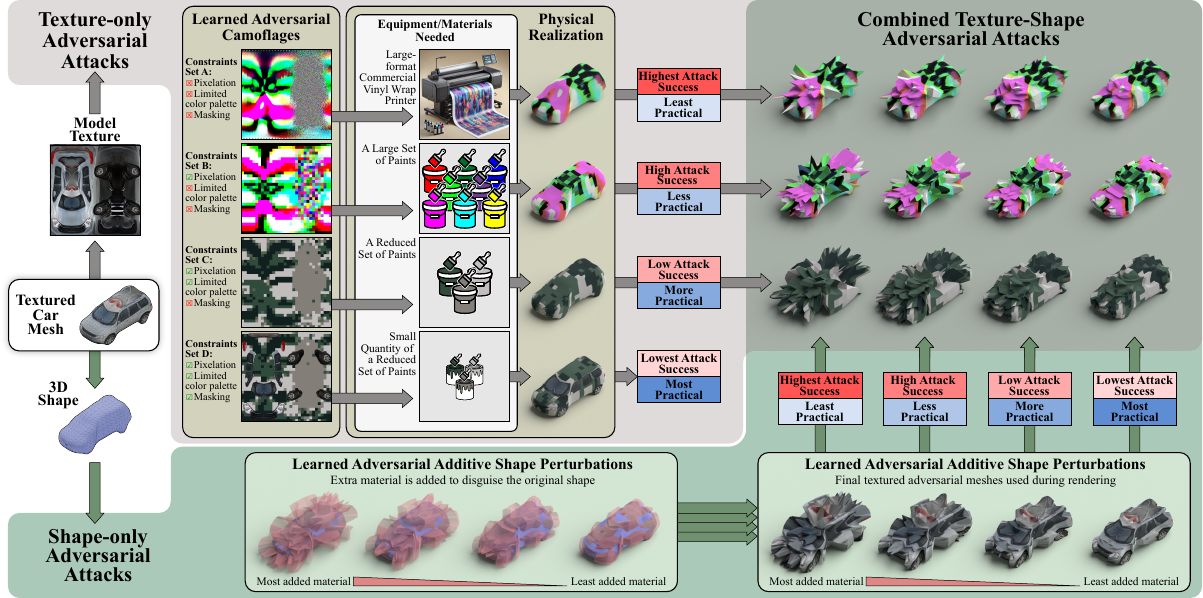}
    \captionof{figure}{We enforced various constraints on common adversarial attacks to improve their implementation in the real world.
        Our pipeline perturbs objects' texture, shape, or both. 
        The latter demonstrates superior efficacy in deceiving object detectors. }
    \label{fig:main_figure}
\end{center}%
}]

\begin{abstract}

Detecting vehicles in aerial images is difficult due to complex backgrounds, small object sizes, shadows, and occlusions. Although recent deep learning advancements have improved object detection, these models remain susceptible to adversarial attacks (AAs), challenging their reliability. Traditional AA strategies often ignore practical implementation constraints. Our work proposes realistic and practical constraints on texture (lowering resolution, limiting modified areas, and color ranges) and analyzes the impact of shape modifications on attack performance. We conducted extensive experiments with three object detector architectures, demonstrating the performance-practicality trade-off: more practical modifications tend to be less effective, and vice versa. We release both code and data to support reproducibility at \url{https://github.com/humansensinglab/texture-shape-adversarial-attacks}.
% The code will be made public upon paper acceptance.

\end{abstract}

%

% Main paper
\section{Introduction}
\label{sec:intro}

Robust object detection in aerial and satellite images is vital for automating critical tasks such as traffic management, urban planning, and disaster response. State-of-the-art detectors, such as YOLO \cite{yolov5} and RetinaNet \cite{RetinaNetPaper}, which are based on deep neural networks (DNN), have become foundational in this domain. However, recent studies such as Szegedy \etal \cite{szegedy2013intriguing} have revealed that DNNs can be susceptible to adversarial examples.  Given the importance of these applications, understanding this vulnerability is crucial, especially in object detection in Remote Sensing Imagery (RSI). Furthermore, there are scenarios where utilizing AAs to impede vehicle detection by computer vision systems in overhead images could offer strategic advantages, such as military camouflage. 

Our primary objective is to investigate the resilience of object detectors against adversarial vehicles in RSI scenarios under realistic constraints. Traditional AA strategies often neglect the physical implementation constraints, focusing solely on task performance. For example, adversarial texture patterns typically resemble those depicted in Figure~\ref{fig:main_figure} (\textit{Texture-only attacks - Constraints Set A}). 
% However, it is essential to consider how these patterns could be realistically applied, such as attaching a printed vinyl wrap to a vehicle. 
However, it is essential to consider how such complex patterns can be produced practically in the physical world. Their creation often requires specialized equipment, such as expensive vinyl wrap printers, and trained professionals to install them on vehicle surfaces.

% We define a \textit{practical adversarial mesh} as a mesh that is perturbed from its original state (by modifying texture and/or shape) such that replicating the same set of modifications in real life would not require specialized equipment or significant resources, or at least would facilitate the process of implementing these modifications.
% We outline the following aspects that define practicality: installation cost, installation difficulty, and difficulty of operation. Our results indicate that while constrained attacks are less successful than traditional AAs~\cite{wang2022fca,Suryanto_2023_ICCV,Suryanto_2022_CVPR,Lian2022BenchmarkingAP}, they provide much better practicality of implementation.  Shape-only attacks exhibit lower effectiveness than unconstrained texture attacks. Nevertheless, combining constrained texture with practical shape modifications enhances performance, reaching levels comparable to unconstrained texture attacks, see Figure~\ref{fig:main_figure}.
In this study, we propose a set of constraints to ensure attacks are feasible for implementation on vehicles, effectively camouflaging them.
We define a \textit{practical adversarial mesh} as a mesh modified in texture and shape such that replicating the modifications in real life requires minimal resources or specialized equipment. We consider practicality based on installation cost, difficulty, and operation. While constrained attacks are less effective than traditional 
% AAs~\cite{wang2022fca,Suryanto_2023_ICCV,Suryanto_2022_CVPR,Lian2022BenchmarkingAP}, 
% AAs~\cite{wang2022fca,Suryanto_2023_ICCV}, 
AAs~\cite{Suryanto_2023_ICCV}, 
they offer better practicality. Shape-only attacks are less effective than unconstrained texture attacks, but combining constrained texture with shape modifications improves performance, reaching levels similar to unconstrained texture attacks (Figure~\ref{fig:main_figure}).

Our work contributes in several ways.
(1) We introduce constrained AAs for shape and texture, designed to create practical 3D camouflages 
% \stan{Probably, there should be a couple of words related to the deformations too?} 
% \mika{camouflages - 3D camouflages. How does that sound?}
capable of deceiving object detectors in RSI. These constraints facilitate a more straightforward implementation compared to unconstrained camouflages.
(2) We thoroughly examine how practicality and adversarial performance relate, finding that they have an inverse relationship.
% (3) We provide a new object detection dataset comprising labeled real overhead images featuring eight distinct vehicle classes covering a broad spectrum of commonly encountered categories. 
(3) We developed a tool for generating synthetic overhead images, contributing to the creation of synthetic datasets.
% \begin{figure}
%     \centering
%     \includegraphics[width=\columnwidth]{Figures/AAs_Comparison.pdf}
%     \caption{
%         Illustration of the three types of adversarial attacks. 
%         %CG --- computer graphics. PBR --- physics-based renderer.
%     }
%     \label{fig:AAs_comparison}
% \end{figure}

\section{Related Work}
\label{sec:rel_work}

Adversarial attacks (AAs) have become central to computer vision research. Szegedy \etal \cite{szegedy2013intriguing} introduced AAs to expose vulnerabilities in deep learning models. Research has since focused on generating adversarial examples 
% \cite{moosavi2016deepfool} 
and divided AAs into digital and physical categories \cite{aa-overview-kazmi}. Digital AAs modify image pixels imperceptibly \cite{du2022adversarial}, while physical AAs manipulate objects in the physical world \cite{Eykholt_2018_CVPR}. A hybrid approach, \textit{simulated AAs}, tests perceptible attacks in simulated environments \cite{kurakin2016adversarial}. Our work aligns with simulated AAs, using synthetic data and realistic physics-based rendering for testing.

% Recent years have seen increased attention by the research community to \textit{remote sensing imagery} 
% \cite{zhu2021tph,YUAN2021114417,rs12030575,s20226485,Xia_2018_CVPR,EIKVIL200965,abady2022manipulation,albert2017using}. These images represent a top-down view and typically have resolutions ranging from several centimeters to several meters or even tens of meters per pixel. 
% The exponential growth of this type of data necessitates robust automated solutions, which directly motivate adversarial attacks --- a tool for identifying vulnerabilities in deep learning models. 
% Although the research community has mostly paid attention to ground-level AAs due to their relevance to the safety-critical field of autonomous driving, there is a list of notable works in the field of RSI AAs 
% \cite{du2022physical,adhikari2020adversarial,wise2022developing,czaja2018adversarial,chen2019adversarial,Yin2022UniversalAP,9726211,rs14215298,rs13204078,10092201,Lian2022BenchmarkingAP}. 
% However, most of these attacks are physically challenging to implement. To our knowledge, the only study implementing an aerial view adversarial attack is \cite{du2022physical}. 
% They employed diverse configurations of adversarial patches, such as placing them on the vehicle or its surroundings, to target vehicle detectors in images. Our approaches differ in the types of constraints applied to the adversarial patches (camouflages in our case) and their stringency.

Recent studies also explore adversarial 3D geometry, mainly in autonomous driving using point clouds and LIDARs \cite{XU2022102122, lee2020shapeadv}. Unlike those works, our focus is solely on RGB data in remote sensing imagery (RSI), where adversarial attacks are underexplored \cite{du2022physical}.
In RSI, the demand for robust automation has risen \cite{zhu2021tph}, driving research on AAs. Many attacks are impractical, so we focus on realistic adversarial camouflages with real-world constraints.
% To the best of our knowledge, the only study addressing physical aerial adversarial attacks is \cite{du2022physical} 
% \stan{Is this still true? [9] is a 2022 paper. Something may have changed in the meantime.}
One of the few studies addressing physical aerial adversarial attacks is \cite{du2022physical}, which uses adversarial patches to target vehicle detectors. Our approach differs in applying stricter constraints to adversarial camouflages and vehicle modifications, while applying these modifications to the entire vehicle and at a lower geo-spatial resolution.
Additionally, instead of real-world tests, we evaluated our camouflages on highly realistic synthetic data produced by a physics-based renderer.

\begin{figure}
    \centering
    \includegraphics[width=\columnwidth]{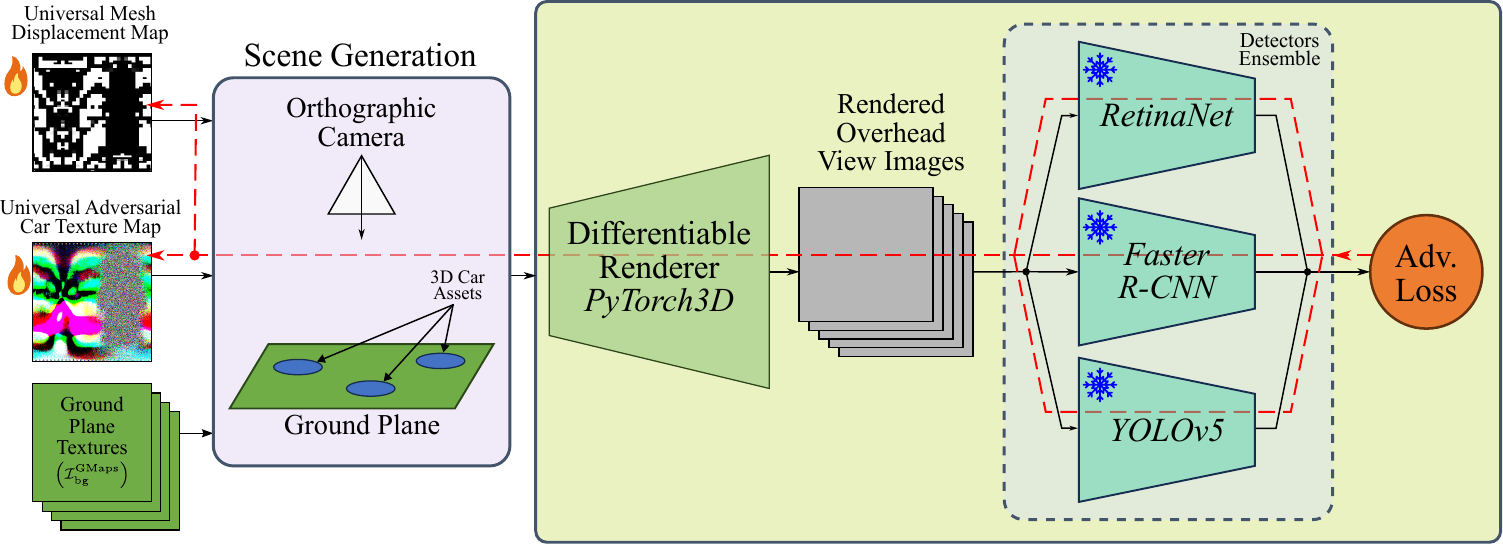}
    \caption{Our pipeline for adversarial attacks on an ensemble of object detectors. The back-propagation path is illustrated by the red dashed line. During inference, we evaluate each model independently.
    % we make predictions using only one model. 
    % We then average the resulting EASRs for all models to analyze the results. Inference with PT3D data follows the same diagram, but instead of using Blender, we use PT3D to render the inference images.
    }
    \label{fig:pipeline}
\end{figure}

\begin{figure}
    \centering
    \includegraphics[width=0.95\columnwidth]{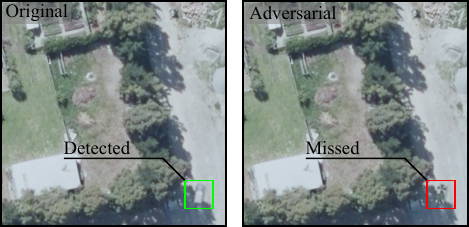}
    \caption{
    % Examples of adversarial images are shown, with the top row displaying the original images and the bottom row showing the corresponding adversarialimages produced using Blender.
    Left: original image, right: corresponding adversarial image generated with Blender.
    % \stan{Needs an improvement}
    }
    \label{fig:example_adversarial_pairs}
\end{figure}
\section{Method}
\label{sec:method}

% \begin{figure*}
%     \centering
%     \includegraphics[width=\textwidth]{Figures/02_FlowChart-tw.pdf}
%     \caption{Our pipeline for adversarial attacks on an ensemble of object detectors. The back-propagation path is illustrated by the red dashed line. During inference, we make predictions using only one model. We then average the resulting EASRs for all models to analyze the results. Inference with PT3D data follows the same diagram, but instead of using Blender, we use PT3D to render the inference images.}
%     \label{fig:pipeline}
% \end{figure*}

% This section outlines the methodology for crafting adversarial vehicles. We provide an overview of the general pipeline employed for generating adversarial vehicles in Section \ref{sec:method:general}. We delve into the specific techniques used to create adversarial textures and shapes in Sections \ref{sec:method:texture} and \ref{sec:method:mesh}, respectively. Section \ref{sec:method:combined} elaborates on the procedure for simultaneously optimizing the texture and shape descriptors of a mesh.
% This section outlines the methodology for generating adversarial vehicles. We describe the general pipeline in Section \ref{sec:method:general}, followed by specific techniques for creating adversarial textures and shapes in Sections \ref{sec:method:texture} and \ref{sec:method:mesh}. Section \ref{sec:method:combined} elaborates on the combined optimization of textures and shapes.

%%%%% PT3D DATA %%%%%
\subsection{PyTorch3D Data Generation}
\label{sec:datasets:synthetic:pt3d}

PyTorch3D (PT3D) data is generated using PyTorch3D \cite{pytorch3d} with 3D vehicle meshes from a GAN-based generator. We adapt and retrain the Textured 3D GAN (T3GAN) \cite{textured3dgan} to enable semantic segmentation map sampling, producing a set of meshes $\mathcal{M}$.
Using GMaps backgrounds $\mathcal{I}_\text{bg}^\text{GMaps}$ and vehicle meshes $\mathcal{M}$, the differentiable renderer (DR) generates images $I = \text{DR}\left(I_\text{bg}^\text{GMaps},M\right)$, where $M$ includes texture $T$ and shape $S$. Post-processing steps like blurring and anti-aliasing enhance realism. $M$ can include either original or adversarial meshes, producing ``original'' or ``adversarial'' images, respectively. The pipeline supports synthetic dataset creation and ground-truth annotations, with adversarial attacks using optimized $S_\text{adv}$ or $T_\text{adv}$.
We use this data to train vehicle detectors and to optimize adversarial attacks.

\subsection{Adversarial Optimization Pipeline}
\label{sec:method:general}

% During each cycle of the attack, we use PT3D to generate a batch of adversarial images $I_b=\{I_1,\dots,I_{N_b}\}$ of size $N_b$, such that
% $
% I_k = \text{DR}\left(I_{\text{bg},k}^\text{GMaps},M_k\right),
% \forall\: k\in[1,\dots,N_b],
% $
% where $I_k$ is the $k$-th image in $I_b$, $I_{\text{bg},k}^\text{GMaps}$ is the $k$-th image from the batch of background images $I_\text{bg}^\text{GMaps}$ sampled from the GMaps dataset $\mathcal{I}_\text{bg}^\text{GMaps}$, $M_k$ is a randomly sampled mesh from $\mathcal{M}$ with its corresponding shape and texture components $S_k$ and $T_k$ respectively, and $\text{DR}$ is a differentiable renderer as described in \Cref{sec:datasets:synthetic:pt3d}. 
% Depending on the optimized entity, either the most recent $S_\text{adv}$ or $T_\text{adv}$ replaces the corresponding counterpart in $M_k$ at the beginning of each iteration. See an overview of our pipeline in \Cref{fig:pipeline}.
Each cycle of the attack uses PT3D to generate a batch of adversarial images $I_b=\{I_1,\dots,I_{N_b}\}$ such that 
$
I_k = \text{DR}\left(I_{\text{bg},k}^\text{GMaps},M_k\right),
\forall\: k\in[1,\dots,N_b],
$
where $I_k$ is the $k$-th image from the batch, $I_{\text{bg},k}^\text{GMaps}$ is the $k$-th background image sampled from the GMaps dataset, $M_k$ is a randomly selected mesh with shape and texture components $S_k$ and $T_k$, and $\text{DR}$ is a differentiable renderer following \Cref{sec:datasets:synthetic:pt3d}. Depending on the optimized entity, either the most recent $S_\text{adv}$ or $T_\text{adv}$ replaces the corresponding counterpart in $M_k$ at the beginning of each iteration.
During the attack, we ensure each image contains only one vehicle to avoid producing meshes that rely on multiple camouflages being in close proximity. The attack aims to create independently effective adversarial meshes.

% During the attack, we ensure that each image contains only one vehicle. We adopt this limitation intentionally to prevent the production of meshes that may rely on multiple camouflaged vehicles being in close proximity to each other. 
% Instead, we aim to produce adversarial meshes that are independently effective. 

% Let $F_i$ be the objective function used to train a detector model $D_i$. We supply the batch of images $I_b$ to $D_i$, producing a set of predictions $y_\text{pred}=D_i\left(I_b\right)$. 
% We then minimize a weighted loss function for an ensemble of synthetic models:
% \begin{gather}
%     \mathcal{L} 
%     = 
%     \label{eq:ensemble-adv-texture-full}
%     \displaystyle\sum_i \lambda_i 
%     \mathbb{E}\left[F_i\left(D_i\left(I_b\left(I_\text{bg}^\text{GMaps},M\right)\right),y_\text{gt}\right)\right], \\
%     M_\text{adv}^\star 
%     =
%     \label{eq:ensemble-adv-textures}
%     \underset{M}{\arg\min} \mathcal{L}\left(M\right),
% \end{gather}
% where $y_\text{gt}$ is the set of ground-truth locations of objects in the given image and is manually replaced with $y_\text{gt}=\emptyset$. $D_i$ represents the $i$-th model from an ensemble of detection models trained on synthetic data. 
% Depending on the optimized entity, either the $S$ or the $T$ component of $M$ is optimizable.
% For each experiment, model coefficients $\lambda_i$ are selected such that the initial loss values $F_i$ for each model fall within the same order of magnitude. We recognize that there are numerous ways to select coefficients. However, we focus only on this one in our study.

Let $F_i$ be the objective function used to train a detector model $D_i$. We supply $I_b$ to $D_i$, producing predictions $y_\text{pred} = D_i(I_b)$. We then minimize a weighted loss function for an ensemble of models:
$
    \mathcal{L} = \sum_i \lambda_i \mathbb{E}\left[F_i(D_i(I_b), y_\text{gt})\right],
    M_\text{adv}^\star = {\arg\min}_M \mathcal{L}(M),
$
where $y_\text{gt}$ are the ground-truth object locations, manually set to $\emptyset$ for adversarial attack training. We use coefficients $\lambda_i$ such that the initial loss values $F_i$ are in the same order of magnitude.

\subsection{Texture-based Attacks}
\label{sec:method:texture}

%%%%% CONSTRAINTS %%%%%
% In texture-based attacks, only one universal texture map is optimized, \ie an adversarial texture map applied to all the meshes.
% While \underline{u}nconstrained adversarial textures (abbreviated as ``U'') achieve excellent performance in fooling vehicle detectors, they are far from being practical for implementation.
% We introduce constraints imposed on textures that allow the production of practical camouflages by reflecting limitations associated with the real world implementation.
% These constraints include the following: 
% \textit{Spatial Resolution}, \textit{Spatial Restriction}, and \textit{Color Restriction}. 
In texture-based attacks, we optimize a universal texture map applied to all meshes. While (\textbf{u})nconstrained adversarial textures (abbreviated as ``U'') achieve excellent performance, they are impractical for real-world use. We introduce constraints to reflect practical implementation limitations: \textit{Spatial Resolution}, \textit{Spatial Restriction}, and \textit{Color Restriction}.

% \textbf{Spatial Resolution.}
% Applying iridescent patterns to irregular shapes such as vehicle surfaces is often challenging \cite{adhikari2020adversarial,Huang_2020_CVPR,wang2022fca,wang2021dual,Suryanto_2023_ICCV,Suryanto_2022_CVPR,du2022physical}. 
% Hence, we adopt the first constraint expressed as texture \underline{pix}elization (abbreviated as ``Pix'') with block sizes of \qtyproduct{16x16}{px}, corresponding to approximately \qtyproduct{15x15}{\cm} on the rooftop of a vehicle. Given that RSI typically does not offer resolutions beyond \qtyproduct{15}{cm/px}, this constraint does not noticeably impact the vehicle's image appearance, significantly streamlining the implementation process. We implement this constraint by storing a latent representation of the adversarial texture as a $32\times32\times3$ tensor. Upon texture generation request, we upscale this tensor to $512\times512\times3$ using the nearest-neighbor interpolation, resulting in a pixelated output.
\textbf{Spatial Resolution.} Applying iridescent patterns to irregular shapes such as vehicle surfaces is often challenging. We impose a texture (\textbf{pix})elization (abbreviated as ``Pix'') constraint with block sizes of $16 \times 16$ px to ensure practical resolution, corresponding to approximately 15 cm on vehicle rooftops. We implement this by storing the adversarial texture as a $32 \times 32 \times 3$ tensor, then upscaling it to the original size of $512 \times 512 \times 3$ via nearest-neighbor interpolation.

% \textbf{Spatial Restriction.}
% Another notable limitation is the necessity for vehicle camouflage not to hinder road visibility. Therefore, specific vehicle areas like glass must remain free of camouflage, which we address by segmenting the meshes used for the attacks into semantic parts. We then apply \underline{ma}sks (abbreviated as ``Ma'') to restrict the area where the adversarial texture is applied. Hence, the second constraint is spatial restriction. It balances reduced performance from a smaller camouflage coverage area with an enhanced operational experience. To implement this constraint, we use an adversarial texture map $T_\text{adv}$, an original texture map $T_\text{or}$ of a vehicle, and a corresponding binary segmentation mask $T_\text{mask}$. Using these three entities, we produce the segmented adversarial texture map as $T_\text{segmented} = T_\text{or}\cdot\left(1 - T_\text{mask}\right) + T_\text{adv}\cdot T_\text{mask}$.
\textbf{Spatial Restriction.} Another notable limitation is the need for vehicle camouflage to not hinder vehicle operation. We restrict the camouflage to specific areas of the vehicle using segmentation (\textbf{ma})sks (abbreviated as ``Ma''). Certain parts, such as windows, remain free from camouflage. The segmented adversarial texture map is given by 
% \stan{Wouldn't it be better if this is abbreviated as $T_\text{seg}$}
% $T_\text{segmented} = T_\text{or} \cdot (1 - T_\text{mask}) + T_\text{adv} \cdot T_\text{mask},$
$T_\text{seg} = T_\text{or} \cdot (1 - T_\text{mask}) + T_\text{adv} \cdot T_\text{mask},$
and $T_\text{or}$ is the original texture.

\textbf{Color Restriction.}
% While the previous two limitations deal with placing textures on the vehicle surface, the final and strictest constraint remains color limitations. 
% We leverage this constraint in two ways: 1) fixing the color count and enabling the attacker to \underline{l}earn optimal \underline{c}olors (abbreviated as ``Lc'') and their placement in the adversarial texture map, or 2) \underline{f}ixing the \underline{c}olors (abbreviated as ``Fc'') and letting the attacker learn their placement in the texture map.
% This constraint stands apart from softer constraints used in previous works, such as \textit{non-printability score}, as it imposes a strict limitation on the number of colors in the output. It is crucial to recognize that while certain constraints mentioned earlier have been investigated in prior studies \cite{Suryanto_2023_ICCV,Suryanto_2022_CVPR,du2022physical,wang2022fca}, the concept of a ``color restriction'' remains largely unexplored in the literature.
% Furthermore, previous research efforts have not adequately addressed the simultaneous integration of the abovementioned constraints.
Our strictest constraint limits the number of colors in the adversarial texture map, implemented in two ways: 1) fixing the color count and (\textbf{l})earning both the (\textbf{c})olors and their placement (``Lc''), or 2) (\textbf{f})ixing the (\textbf{c})olors and optimizing their placement only (``Fc''). Unlike softer constraints such as the \textit{non-printability score}, this enforces strict color limits, a concept largely unexplored in prior works.
To enforce this constraint, the color of each pixel is determined during optimization by predicting a probability distribution $p(c)$ over a fixed set of colors. This distribution is sharpened using a double softmax $s(\cdot)$ to amplify the most probable color: $p_A(c) = s(s(p(c)))$. The pixel color is initially set to $\mathbb{E}[p_A(c)]$, approximating the mode color. After optimization, each pixel is assigned $\arg\max_{c_i}p(c_i)$, producing a camouflage that satisfies the color constraint. Additional details can be found in \Cref{sec:SM:practicality_and_comparisons:texture-based_attacks} in the Supplementary Material.

% Finding an adversarial texture map with constrained colors involves determining the color of each pixel. At each attack iteration, we predict a probability distribution $p(c)$ for each pixel over a set of colors $c$. We then amplify the most probable color while damping the others using a softmax function $s(\cdot)$ twice: $p_A(c)=s(s(p(c)))$. The pixel color is set to $\mathbb{E}\left[p_A(c)\right]$, which is close to the mode color due to amplification. After optimization, each pixel is assigned $\arg\max_{c_i}p(c_i)$, resulting in a camouflage that meets the color constraint and is similar to the optimized camouflage which uses $\mathbb{E}\left[\cdot\right]$ instead of $\arg\max$. For more details on our constraints, please refer to the supplementary material.

\subsection{Shape-based Attacks}
\label{sec:method:mesh}

We optimize a universal perturbation in shape-based attacks using a 2D displacement map from a common UV map. Deformations extend outward from the mesh center, preserving the original shape. We enforce \textit{Symmetry} for balanced mass and \textit{Perturbation Magnitude} (PM) to limit deformation.
% We introduce two constraints: \textit{Symmetry} and \textit{Perturbation Magnitude} (PM). The symmetry constraint ensures balanced mass distribution, implemented by introducing a symmetry axis for mirroring the displacement map. The PM constraint restricts deformation volume.

% \begin{figure}
%     \centering
%     \includegraphics[width=0.9\columnwidth]{Figures/Adversarial_Pair_Examples.pdf}
%     \caption{
%     % Examples of adversarial images are shown, with the top row displaying the original images and the bottom row showing the corresponding adversarialimages produced using Blender.
%     Top row shows original images, bottom row shows corresponding adversarial images generated with Blender.
%     }
%     \label{fig:example_adversarial_pairs}
% \end{figure}

\subsection{Combined Attacks}
\label{sec:method:combined}
\label{sec:method:combined:sequential}
\label{sec:method:combined:parallel}

% We also conduct combined adversarial attacks, which result in vehicles with adversarial texture and shape. We split them into two branches: \textit{sequential} and \textit{parallel} depending on how the attacks are performed. 
We also conduct combined attacks where both texture and shape are optimized. These can be performed sequentially or in parallel.
In \textit{sequential combined attacks}, we first optimize the texture map and then perform a shape-based attack using the adversarial texture. This allows us to evaluate the performance of both attacks sequentially.
In \textit{parallel combined attacks}, we alternate between optimizing the texture and shape. Each entity is optimized for a fixed number of steps $n_\text{pll}$, switching between them until the loss converges.

% In \textit{sequential combined adversarial attacks}, we borrow an adversarial texture map obtained from a texture-based attack, use it instead of vehicle texture, and perform a shape-based attack to optimize the displacement map described in \Cref{sec:method:mesh}. Thus, both mesh properties are attacked sequentially. The difference between this and the shape-based attacks is the initial texture map: instead of the original textures, an adversarial texture is utilized. 
% In addition, similar to shape-based attacks, we run a series of experiments to establish the optimal $\text{PM}$ for each sequential attack.

% In \textit{parallel combined adversarial attacks}, both texture and shape are optimized. 
% We alternate between optimizing the adversarial texture map and the displacement map: optimizing one entity for a fixed number of steps $n_\text{pll}$, then switching to the other entity for another $n_\text{pll}$ steps. 
% We continue this process until the loss converges.

\begin{table}
    \centering
    \footnotesize
    \caption{
    % Comparing the practicality of the attacks explored in our study to previous works. We do not distinguish between sequential and parallel combined attacks as they only impact the process, not the final result's form. The first symbol reflects the texture-related practicality score, the second number reflects the shape-related practicality score.
    Comparison of attack practicality with prior works. \textcolor{blue}{Texture practicality} is the first score, \textcolor{orange}{shape practicality} is the second. Sequential and parallel combined attacks yield identical final results. Full table is in \Cref{sec:SM:practicality_and_comparisons} in the Supplementary Material.
    }
    \begin{tabularx}{\columnwidth} { 
      >{\centering\arraybackslash}m{3mm}| 
      >{\raggedright\arraybackslash}m{2.3cm}| 
      >{\centering\arraybackslash}m{5mm}| 
      >{\centering\arraybackslash}m{5mm}| 
      >{\centering\arraybackslash}m{5mm}| 
      >{\centering\arraybackslash}m{7mm} | 
      >{\centering\arraybackslash}X
      }
     & \textbf{Camouflage} & \textbf{PC} & \textbf{DI} & \textbf{DO} & \textbf{Score} & \textbf{Notes} \\
     \hline
     \multirow{3}{*}{\rotatebox{90}{Other}} 
     & Du \etal (ON) \cite{du2022physical}  & $\textcolor{blue}{+}\textcolor{orange}{0}$ & $\textcolor{blue}{+}\textcolor{orange}{0}$ & $\textcolor{blue}{+}\textcolor{orange}{0}$ & 3 & Small AA area \\
     \cline{2-7}
     & Du \etal (OFF) \cite{du2022physical} & $\textcolor{blue}{0}\textcolor{orange}{0}$ & $\textcolor{blue}{+}\textcolor{orange}{0}$ & $\textcolor{blue}{-}\textcolor{orange}{0}$ & $0$ & Limited mobility \\
     \cline{2-7}
     % & EVD4UAV \cite{sun2024evd4uav} & $+0$ & $+0$ & $+0$ & $+3$ & Small AA area \\
     % \cline{2-7}
     % & FCA \cite{wang2022fca}               & $-0$ & $-0$ & $+0$ & $-1$ & Special equipment \\
     % \cline{2-7} 
     % & ACTIVE \cite{Suryanto_2023_ICCV}     & $-0$ & $-0$ & $+0$ & $-1$ & Special equipment \\
     % \cline{2-7}
     & DTA \cite{Suryanto_2022_CVPR}        & $\textcolor{blue}{-}\textcolor{orange}{0}$ & $\textcolor{blue}{-}\textcolor{orange}{0}$ & $\textcolor{blue}{+}\textcolor{orange}{0}$ & $-1$ & Special equipment \\
     \hline
     \multirow{5}{*}{\rotatebox{90}{Our}} 
     & T-U          & $\textcolor{blue}{-}\textcolor{orange}{0}$ & $\textcolor{blue}{-}\textcolor{orange}{0}$ & $\textcolor{blue}{-}\textcolor{orange}{0}$ & $-3$ & Special equipment \\
     \cline{2-7}
     & T-Ma         & $\textcolor{blue}{-}\textcolor{orange}{0}$ & $\textcolor{blue}{-}\textcolor{orange}{0}$ & $\textcolor{blue}{+}\textcolor{orange}{0}$ & $-1$ & Special equipment \\
     % \cline{2-7}
     % & T-PixMa      & $-0$ & $+0$ & $+0$ & $+1$ & Stickers/paint squares \\
     \cline{2-7}
     & T-PixFcMa    & $\textcolor{blue}{+}\textcolor{orange}{0}$ & $\textcolor{blue}{+}\textcolor{orange}{0}$ & $\textcolor{blue}{+}\textcolor{orange}{0}$ & $+3$ & Lim. color stickers \\
     \cline{2-7}
     & S-O          & $\textcolor{blue}{0}\textcolor{orange}{-}$ & $\textcolor{blue}{0}\textcolor{orange}{-}$ & $\textcolor{blue}{0}\textcolor{orange}{-}$ & $-3$ & Shape modification \\
     \cline{2-7}
     & C-Fc         & $\textcolor{blue}{-}\textcolor{orange}{-}$ & $\textcolor{blue}{-}\textcolor{orange}{-}$ & $\textcolor{blue}{-}\textcolor{orange}{-}$ & $-6$ & Spec. eq./shape mod.\\
     % \cline{2-7}
     % & C-PixFc      & $+-$ & $+-$ & $--$ & $-2$ & Shape modification \\
     \hline
    \end{tabularx}
    \label{tab:comparisons}
\end{table}

\subsection{Computational Requirements}

% The pipeline involves computationally intensive rendering and optimization. 
Experiments ran on a machine with an Intel Xeon Gold 6252 CPU, 755 GB RAM, and an NVIDIA Quadro RTX 6000 GPU (24 GB). Each attack took ~3 hours per epoch, using up to 5 GB RAM and 12 GB GPU memory. 
% Future work may explore parallel batch processing for greater efficiency.
\section{{Practicality and Comparisons}}
\label{sec:comparisons}

Our focus is not on enhancing AAs performance but exploring the impact of realistic constraints. Given the high effectiveness of unconstrained AAs, there is limited room for improvement. We evaluate our work based on three qualitative criteria: \textit{production cost} (PC), \textit{difficulty of installation} (DI), and \textit{difficulty of operation} (DO), rated as good ($+$), insignificant ($0$), or bad ($-$). Practical camouflages score higher. 
See definitions in \Cref{sec:SM:practicality_and_comparisons} in the Supplementary Material.
% Definitions are in the Supplementary Material.

% Based on the outlined criteria and as depicted in \Cref{tab:comparisons}, the most practical camouflage is the texture-based T-PixFcMa, despite having one of the lowest EASR (\Cref{sec:experiments}). The only works that have comparable practicality scores are Du \etal \cite{du2022physical} (ON) and EVD4UAV \cite{sun2024evd4uav}. However, the areas occupied by their patches are very small for being effective in aerial images of the resolution considered in our work.
% Please refer to the supplementary material for a more in-depth description of the score assignment.
% Results from \Cref{tab:comparisons,tab:texture_results_PT3D,tab:pert_per_texture} illustrate the trade-off between practicality and performance. While some previous studies excel in AAs for vehicle detectors, they lack practicality. We conclude that performance and practicality are inversely related when optimization for performance is the goal. For example, for randomly generated camouflages (\Cref{tab:texture_results_PT3D}), the reduction in performance may not be justifiable because no optimization is carried out.
% As shown in \Cref{tab:comparisons}, the most practical camouflage is T-PixFcMa, despite its reduced effectiveness (\Cref{tab:texture_results_PT3D}). 
As shown in \Cref{tab:comparisons}, T-PixFcMa is the most practical camouflage, despite lower effectiveness (\Cref{tab:texture_results_PT3D}).
% Comparable practicality scores are seen in Du \etal \cite{du2022physical} (ON) and EVD4UAV \cite{sun2024evd4uav}, but their patch sizes are too small for effective use in aerial images at the resolution in our work. 
Du \etal \cite{du2022physical} (ON) and EVD4UAV \cite{sun2024evd4uav} achieve similar practicality scores, but their patches are too small for effective use in aerial imagery at our resolution.
More details on score assignment are in \Cref{sec:SM:practicality_and_comparisons} in the Supplementary Material. Results from \Cref{tab:comparisons,tab:texture_results_PT3D,tab:pert_per_texture} highlight the trade-off between practicality and performance. While some studies excel in AA performance, they lack practicality. We conclude that optimizing for performance reduces practicality. For example, randomly generated camouflages (\Cref{tab:texture_results_PT3D}) show performance reduction that may not be justified without optimization.
While a more rigorous analysis incorporating real data could be done (e.g., user studies for DO), this is beyond the scope due to limited resources.

\begin{table}
        \centering
        \footnotesize
        \caption{
            % The figures show mean values from evaluations of individual synthetic models on PT3D and Blender data. ``T'', ``R'', ``S'' and ``C'' represent the texture, random texture, shape, and combined attacks. 
            % Note that Lc and Fc are mutually exclusive by definition. 
            % $\text{PM}^\star$ and $\text{Pr}^\star$ represent the optimal perturbation magnitude and practicality of the attacks involving shape modifications.
            % The figures show mean values from synthetic models evaluated on PT3D and Blender data. ``T'', ``R'', ``S'', and ``C'' denote texture, random texture, shape, and combined attacks. 
            % Lc and Fc are mutually exclusive. 
            % $\text{PM}^\star$ and $\text{Pr}^\star$ indicate optimal perturbation magnitude and practicality for shape attacks. See the full table in the supplementary material.
            Figures show mean values from synthetic models on PT3D and Blender data. ``T'', ``R'', ``S'', and ``C'' represent \textcolor{blue}{texture}, \textcolor{blue}{random texture}, \textcolor{orange}{shape}, and \textcolor{Magenta}{combined} attacks. Lc and Fc are mutually exclusive. $\text{PM}^\star$ and $\text{Pr}^\star$ denote optimal perturbation magnitude and practicality for shape attacks. 
            See \Cref{tab:texture_results_PT3D_SM} in the Supplementary Material for the full table.
            % Full table is in the Supplementary Material.
        }
        \label{tab:texture_results_PT3D}
        \label{tab:pert_per_texture}
        \begin{tabularx}{\columnwidth}{
            >{\raggedright\arraybackslash}m{17mm}| 
            >{\centering\arraybackslash}m{5mm}| 
            >{\centering\arraybackslash}m{5mm}| 
            >{\centering\arraybackslash}m{5mm}| 
            >{\centering\arraybackslash}m{5mm}| 
            >{\centering\arraybackslash}m{6mm}| 
            >{\centering\arraybackslash}m{6mm}|
            >{\centering\arraybackslash}X| 
            >{\centering\arraybackslash}X
        }
    
            \multirow{2}{*}{\parbox{17mm}{\centering\textbf{Attack}}} & 
            \multicolumn{4}{c|}{\textbf{Constraints}} & 
            $\text{PM}^\star$ &
            $\text{Pr}^\star$ &
            \textbf{PT3D} &
            \textbf{Blender} \\
            \cline{2-5}
            
             & Pix & Lc & Fc & Ma & & &
             \textbf{EASR} & 
             \textbf{EASR} \\
            \hline
            \textcolor{blue}{T-U} & & & &                                         & --- & --- & 95.77\% & 70.02\% \\
            \textcolor{blue}{T-Ma} & & & & \checkmark                             & --- & --- & 75.76\% & 44.43\% \\
            \textcolor{blue}{T-Pix} & \checkmark & & &                            & --- & --- & 94.75\% & 63.83\% \\
            % T-PixMa & \checkmark & & & \checkmark               & --- & --- & 73.39\% & 41.49\% \\
            % T-Lc & & \checkmark & &                             & --- & --- & 92.63\% & 74.65\% \\
            % T-Fc & & & \checkmark &                             & --- & --- & 37.22\% & 55.96\% \\
            % T-LcMa & & \checkmark & & \checkmark                & --- & --- & 71.93\% & 52.51\% \\
            % T-FcMa & & & \checkmark & \checkmark                & --- & --- & 12.58\% & 26.43\% \\
            % T-PixLc & \checkmark & \checkmark & &               & --- & --- & 96.73\% & 68.85\% \\
            % T-PixFc & \checkmark & & \checkmark &               & --- & --- & 35.68\% & 55.87\% \\
            \textcolor{blue}{T-PixLcMa} & \checkmark & \checkmark & & \checkmark  & --- & --- & 68.39\% & 42.15\% \\
            \textcolor{blue}{T-PixFcMa} & \checkmark & & \checkmark & \checkmark  & --- & --- & 12.70\% & 44.64\% \\
            \hline
            % R-U & & & &                                         & --- & --- & 0.88\% & 16.31\% \\
            % R-Pix & \checkmark & & &                            & --- & --- & 3.86\% & 18.95\% \\
            % R-Fc & & & \checkmark &                             & --- & --- & 3.30\% & 18.95\% \\
            \textcolor{blue}{R-PixFc} & \checkmark & & \checkmark &               & --- & --- & 3.16\% & 20.24\% \\
            \hline
            \textcolor{orange}{S-O} & --- & --- & --- & ---                         & 0.4 & 0.6 & 89.82\% & 78.86\% \\
            \hline
            % C-U & & & &                                         & 0.0 & 1.0 & 96.51\% & --- \\
            % C-Pix & \checkmark & & &                            & 0.0 & 1.0 & 95.02\% & --- \\
            % C-Lc & & \checkmark & &                             & 0.0 & 1.0 & 92.68\% & --- \\
            \textcolor{Magenta}{C-Fc (seq.)} & & & \checkmark &                      & 0.2 & 0.8 & 86.80\% & 70.37\% \\
            \textcolor{Magenta}{C-Fc (par.)} & & & \checkmark &                      & 0.2 & 0.8 & 87.11\% & 68.07\% \\
            % C-PixLc & \checkmark & \checkmark & &               & 0.0 & 1.0 & 96.40\% & --- \\
            \textcolor{Magenta}{C-PixFc (seq.)} & \checkmark & & \checkmark &        & 0.2 & 0.8 & 86.83\% & 75.76\% \\
            \textcolor{Magenta}{C-PixFc (par.)} & \checkmark & & \checkmark &        & 0.2 & 0.8 & 89.34\% & 77.86\% \\
    \end{tabularx}
\end{table}

% We recognize that this analysis could be carried out more rigorously and incorporate numerical comparisons from real data. For example, DO could be evaluated through extensive user studies. However, due to limited resources, we leave such analysis beyond the scope of the paper and rely on the assumptions made above.

\begin{figure}
    \centering
    \includegraphics[width=0.8\columnwidth]{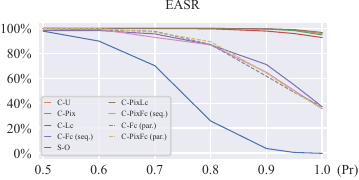}
    \caption{EASR vs Practicality ($\text{Pr}$) curves for the shape-based and combined attacks on PT3D.}
    \label{fig:easr_vs_pr}
\end{figure}

\section{Experiments and Results}
\label{sec:experiments}
\label{sec:method:evaluation}

While \Cref{sec:comparisons} qualitatively compares practicality --- considering production cost, installation, and operational complexity --- this section provides a quantitative evaluation of effectiveness under different adversarial attack scenarios.

\subsection{Evaluation metrics} %Effective Attack Success Rate}
\label{sec:experiments:eval_metrics}
We use effective attack success rate (EASR) to evaluate attack performance, defined as  
$\text{EASR} = \frac{|V_{d,m}| - |V_{m,d}|}{|V_{d,d} \cup V_{d,m}|},$  
where the first subscript in $V_{i,j}$ indicates whether a vehicle was detected ($d$) or missed ($m$) in the original dataset, and the second subscript indicates whether it was detected ($d$) or missed ($m$) in the adversarial dataset after applying adversarial modifications. EASR is more conservative than the traditional ASR. More details can be found in \Cref{sec:supp_AdditionalResults:EASR} in the Supplementary Material.

\subsection{Test Data}
\label{sec:datasets:real}
\label{sec:datasets:real:linz}
\label{sec:datasets:real:gmaps}
\label{sec:datasets:synthetic:blender}

We describe below the datasets used to test the trained models and the camouflages generated from the attacks. 
% All data will be made publicly available upon the paper's acceptance.

%%%%% LINZ DATASET %%%%%
\textbf{LINZ Dataset.}
% \label{sec:datasets:real:linz}
We used the labeled LINZ dataset using LINZ Data Service aerial orthoimages\footnote{\url{https://data.linz.govt.nz/layer/51926-selwyn-0125m-urban-aerial-photos-2012-2013/}}. The dataset was sampled into 384x384px images, resulting in \num{172595} images with a \qty{12.5}{\cm\per px} resolution.
We utilized only the ``Small Vehicles'' class for experiments, removing other labels but preserving all images. This dataset is denoted as $\mathcal{I}^\text{LINZ}$.
A second dataset, $\mathcal{I}_\text{bg}^\text{LINZ}$, was created by removing vehicles from $\mathcal{I}^\text{LINZ}$ using ``Inpaint Anything'' \cite{yu2023inpaint}.

%%%%% GMAPS DATASET %%%%%%
\textbf{GMaps Dataset}
% \label{sec:datasets:real:gmaps}
The GMaps dataset provides background images \(\mathcal{I}_\text{bg}^\text{GMaps}\) for the PyTorch3D \cite{pytorch3d} data generation pipeline (\Cref{sec:datasets:synthetic:pt3d}). We extracted Google Maps (GMaps) satellite images \(\mathcal{I}^\text{GMaps}\) with matching coordinates to the LINZ aerial dataset using QGIS\footnote{\url{https://www.qgis.org}},
% ~\cite{qgis}
ensuring a direct LINZ-GMaps correspondence. Real vehicles were manually removed from these images using an image editor,
%Adobe Photoshop\footnote{\url{https://www.adobe.com/products/photoshop.html}}
% ~\cite{adobephotoshop}
resulting in the background set \(\mathcal{I}_\text{bg}^\text{GMaps}\), later used in the synthetic data generation process.
% (\Cref{sec:datasets:synthetic:pt3d}).

%%%%% BLENDER DATA %%%%%
\textbf{Blender Data Generation}
% \label{sec:datasets:synthetic:blender}
% We use the \textit{Blender data} to test the adversarial meshes in a more realistic scenario. We generate a test set for each adversarial texture or geometry using realistic Blender Cycles \cite{Cycles} physics-based renderer.
% Unlike PyTorch3D, Blender renders much more realistic data. However, because it is non-differentiable, we do not use it in the adversarial optimization process.
% To produce a synthetic overhead image with Blender, we use an image sampled from $\mathcal{I}_\text{bg}^\text{LINZ}$ and a set of meshes $M$, which can consist of either original meshes or meshes utilizing adversarial texture and/or shape.
Blender data tests adversarial meshes in realistic settings using the Cycles physics-based renderer \cite{Cycles}. Unlike PT3D, Blender generates highly realistic but non-differentiable images, excluding it from adversarial optimization. Synthetic overhead images are produced by Blender using LINZ backgrounds $\mathcal{I}_\text{bg}^\text{LINZ}$ and meshes $M$, incorporating either original or adversarial textures and shapes.

%%%%% MODELS %%%%%
\subsection{Detection Models}
\label{sec:method:models}

% We used three model architectures for vehicle center detection in RSI, including RetinaNet \cite{RetinaNetPaper}, Faster R-CNN \cite{fasterrcnn}, and YOLOv5 \cite{yolov5}. We chose these models to represent YOLO-family detectors (YOLOv5), one-stage detectors (RetinaNet), and two-stage detectors (Faster R-CNN). Each architecture was trained on both real and synthetic datasets, resulting in models labeled as ``real'' and ``synthetic'' (abbreviated as ``synt''). On the real test set, synthetic models achieve around $50\%$--$63\%$ average precision, while real models score above $80\%$. Detailed performance figures are available in the Supplementary Material. As expected, models perform better within their training domains. We attacked ensembles of synthetic models and used only one synthetic model for inference.
% We employed RetinaNet \cite{RetinaNetPaper}, Faster R-CNN \cite{fasterrcnn}, and YOLOv5 \cite{yolov5} for vehicle center detection in RSI, representing one-stage, two-stage, and YOLO-family detectors. Each was trained on real and synthetic datasets, labeled as ``real'' and ``synt'' models. On the real test set, synthetic models achieve $50\%$--$63\%$ average precision, while real models exceed $80\%$. Performance details are in the Supplementary Material. We attacked synthetic model ensembles, using one synthetic model for inference.
We used RetinaNet \cite{RetinaNetPaper}, Faster R-CNN \cite{fasterrcnn}, and YOLOv5 \cite{yolov5} for vehicle center detection in RSI, representing one-stage, two-stage, and YOLO-family detectors. Each was trained on real and synthetic data, labeled ``real" and ``synt" models. On real test data, synthetic models achieved $50$--$63\%$ AP, while real models exceeded $80\%$. More details can be found in \Cref{sec:supp_TrainingDetails} in the Supplementary Material. We attacked synthetic model ensembles, using one for inference.

%%%%% INTRO TO TEXTURE %%%%%
\subsection{Texture-based Attacks}

%%%%% PT3D RESULTS %%%%%
% We alter a vehicle's texture in \textit{texture-based} attacks, aiming to hide it from detectors. We initialize the adversarial textures randomly.
% Given the set of constraints described in \Cref{sec:method:texture}, we conclude that there are twelve possible distinct adversarial texture settings (\Cref{tab:texture_results_PT3D}-upper half). In each setting, we attack the same ensemble of three synthetic models (RetinaNet, Faster R-CNN, YOLOv5) to obtain the complete set of adversarial textures. We also generate four random texture maps to compare the effectiveness of the adversarial textures to those of a random guess.
% See each texture setting, denoted as T-*, and the corresponding results on PT3D test data in \Cref{tab:texture_results_PT3D}. We also report the attack results using randomly generated textures, denoted as R-*.
% In Fc experiments, five colors were selected using K-means clustering on background image pixels $I_\text{bg}^\text{GMaps}$. In Lc experiments, the model learned the placement of five colors.
We modify a vehicle's texture in \textit{texture-based attacks} to conceal it from detectors, starting with random adversarial textures. There are twelve distinct texture settings, each evaluated on an ensemble of three synthetic models (RetinaNet, Faster R-CNN, YOLOv5). We compare these adversarial textures with four random texture maps (R-*). Results on PT3D test data are shown in \Cref{tab:texture_results_PT3D}. In Fc experiments, five colors are determined via K-means clustering of background pixels, while Lc experiments involve model-learned color placement.

To account for the distribution gap between real and synthetic datasets, we repeat the experiments using Blender. The results, presented in \Cref{tab:texture_results_PT3D}, show that constraints reduce performance but increase practicality. Example images are in \Cref{fig:example_adversarial_pairs}, with additional examples in the Supplementary Material. The performance-practicality trade-off remains even when testing on a different domain, like Blender.
We also observe that unrestrained color distribution results in saturated colors, a common but underexplored issue in prior works. Further details are in \Cref{sec:supp_TrainingDetails} the Supplementary Material.

% We examine color distribution in adversarial texture maps and find that lacking color constraints leads to saturated colors at RGB cube corners. This issue persists in prior works too.
% See the Supplementary Material for details.

\subsection{Shape-based Attacks}
\label{sec:experiments:mesh}

In \textit{shape-based attacks}, we alter the geometry of the vehicle sacrificing practicality for improved adversarial performance. We link practicality, denoted as $\text{Pr}$, to the perturbation magnitude $\text{PM}$ as $\text{Pr} = 1 - \text{PM}$, where $\text{PM}\in[0,1]$. $\text{Pr}=0$ indicates extreme mesh perturbation and $\text{Pr}=1$ indicates no perturbation for a more realistic scenario.

Our goal is to minimize $\text{PM}$ (maximize $\text{Pr}$) in shape-based attacks while maximizing EASR performance. We assess multiple attacks on synthetic models across a range of $\text{Pr}$ values. Refer to the curve in \Cref{fig:easr_vs_pr} under ``Original'' for details (utilizing original vehicle textures).
Given an EASR-vs-Pr curve, we compute a practicality metric $\text{P1}$ as the harmonic mean of the EASR and $\text{Pr}$ values for each point on the curve, \ie
$\text{P1} = 2\frac{\text{EASR}\cdot\text{Pr}}{\text{EASR}+\text{Pr}}$.
We then pick the attack with the highest $\text{P1}$ as the optimal one. See the results, denoted as S-O in \Cref{tab:pert_per_texture}. The results suggest that when no adversarial texture is utilized along with a deformed vehicle geometry, the deformation must be large to achieve good performance, which is expensive to produce and difficult to install and operate, rendering it impractical.

%%%%% INTRO TO MESH %%%%%
\subsection{Combined Attacks}
\label{sec:experiments:combined}
\label{sec:experiments:combined:sequential}
\label{sec:experiments:combined:parallel}

In the \textit{combined attacks}, we aim to boost the adversarial performance by attacking both mesh entitites: texture and shape. We discard the adversarial textures that use masking because a modified mesh is hard to segment into semantically meaningful parts. Thus, this leaves us with six adversarial texture maps for mesh-based attacks, each with its texture setting.

\textbf{Sequential Attacks.}
We conduct six sequential attacks, each using one of the six adversarial textures from a non-masked setup. We follow the methodology in \Cref{sec:experiments:mesh} to determine the optimal $\text{PM}$. The results, labeled C-* in \Cref{tab:pert_per_texture}, reveal significant insights. While the improvement over texture attacks without the fixed colors constraint (T-U, T-Pix, T-Lc, T-PixLc) is unjustified due to practicality loss, the fixed colors constraint (T-Fc and T-PixFc) justifies sacrificing some practicality for better performance. Compared to the huge practicality loss in shape-based attacks, the sequential blend of adversarial texture and shape is more efficient than shape-only attacks. We evaluate the resulting adversarial meshes on Blender data where the $\text{PM}^\star\neq0$.
% optimal perturbation magnitude is not 0.

\textbf{Parallel Attacks.}
We conduct parallel attacks using only two adversarial texture maps, Fc and PixFc, to reduce the $\text{PM}$ even further than the sequential attacks. The results in \Cref{tab:pert_per_texture} suggest no significant gain in switching to parallel attacks.

\section{Conclusion}
\label{sec:conclusion}
\label{sec:conclusions}

This study outlines a methodology for developing effective camouflage strategies to conceal vehicles in RSI. We also study the performance-practicality trade-off when implementing adversarial camouflages. While our findings could be misused, it is vital for the research community to be aware of the vulnerabilities in current models that we highlight. We show an inverse relationship between practicality and performance: unconstrained adversarial textures are highly effective against vehicle detection systems, while practical constrained textures are easier to implement but less effective. Shape-only attacks are also less impactful than texture attacks, but combining both can achieve results similar to unconstrained textures. Notably, sequential and parallel executions of shape and texture attacks demonstrate similar adversarial performance. Additionally, we present two pipelines for generating synthetic aerial images: using a differentiable renderer and a physics-based renderer.

% \newpage
\section*{Acknowledgements}  
We thank Brent Lance for valuable feedback and support. This work was supported in part by the Army Research Lab.

% References should be produced using the bibtex program from suitable
% BiBTeX files (here: strings, refs, manuals). The IEEEbib.bst bibliography
% style file from IEEE produces unsorted bibliography list.
% -------------------------------------------------------------------------
\bibliographystyle{IEEEbib}
\bibliography{refs}

% Supplementary Material
\clearpage
% \appendix
\setcounter{figure}{0}
\setcounter{table}{0}
\setcounter{section}{0}
\renewcommand{\thefigure}{S\arabic{figure}}
\renewcommand{\thetable}{S\arabic{table}}
\renewcommand{\thesection}{S\arabic{section}}

\twocolumn[{%
\renewcommand\twocolumn[1][]{#1}%
\begin{center}
    {\LARGE \textbf{Texture- and Shape-based Adversarial Attacks for Overhead Image Vehicle Detection}}\\[1.5ex]
    {\large \textbf{Supplementary Material}}\\[2ex]
\end{center}
}]

\section{Training Details \& Detection Models}
\label{sec:supp_TrainingDetails}

\subsection{Training Details}

We use three model architectures in our experiments: RetinaNet, Faster R-CNN, and YOLOv5. We obtain RetinaNet and Faster R-CNN from Detectron2\footnote{\url{https://github.com/facebookresearch/detectron2}}. We retrieve YOLOv5 from its native implementation by Ultralytics\footnote{\url{https://github.com/ultralytics/yolov5}}. We train the first two using the Detectron2 pipeline. We train YOLOv5 using its native pipeline.

\subsubsection{Real Models}
The training data is derived from the LINZ dataset $\mathcal{I^\text{LINZ}}$, by removing all non-``Small Vehicle'' class labels from the training set, resulting in \num[]{119691} training images.
We train the real RetinaNet and Faster R-CNN for \num[]{10000} iterations and batch size \num[]{640}. We train the real YOLOv5 for \num[]{50} epochs and batch size \num[]{640}, which corresponds to approximately \num[]{10000} iterations. 

\subsubsection{Synthetic Models}
To train the synthetic models, we produce a training synthetic dataset using PT3D consisting of \num[]{30000} images. We train RetinaNet and Faster R-CNN for \num[]{10000} iterations using batch size \num[]{128}, while YOLOv5 is trained for \num[]{42} epochs using batch size \num[]{128} (approximately \num[]{9800} iterations).

\subsection{Detection Models}

\Cref{tab:model_performances} shows the average precision scores by the real and synthetic models on the synthetic and real test sets.
These results supplement Section 7.2 in the paper.

\begin{table}[h]
    \centering
    \footnotesize
    \caption{Evaluation results of the six models on the real and synthetic test sets.}
    \begin{tabularx}{\columnwidth} { 
      >{\raggedright\arraybackslash}p{17mm} |
      >{\centering\arraybackslash}p{10mm} |
      >{\centering\arraybackslash}p{13mm} |
      >{\centering\arraybackslash}X |
      >{\centering\arraybackslash}X |
      >{\centering\arraybackslash}X
      }
     \multicolumn{1}{c|}{\textbf{Architecture}} & \textbf{Training Data} & \textbf{Detection Threshold} & \textbf{AP} (real) &  \textbf{AP} (synt.) & \textbf{AP} (Blender) \\
     \hline
     RetinaNet          & $I^\text{LINZ}$   & 49.82\% & 93.50\%  & 94.80\% & 91.22\% \\
     Faster R-CNN       & $I^\text{LINZ}$   & 72.13\% & 80.34\%  & 93.31\% & 81.71\% \\
     YOLOv5             & $I^\text{LINZ}$   & 59.85\% & 96.21\%  & 95.51\% & 95.55\% \\
     \hline
     RetinaNet          & \textit{PT3D}     & 47.64\% & 49.09\%  & 99.88\% & 79.35\% \\
     Faster R-CNN       & \textit{PT3D}     & 86.95\% & 59.21\%  & 99.49\% & 85.50\% \\
     YOLOv5             & \textit{PT3D}     & 60.40\% & 63.54\%  & 99.95\% & 97.99\% \\
    \end{tabularx}
    \label{tab:model_performances}
    \label{tab:original_blender_results}
\end{table}
\section{PyTorch3D Data Reality Gap Mitigation}
\label{sec:supp_PT3DPostProcessing}

% \stan{Probably, there could be some short intro for this section on why the reality gap mitigation is important.}

Mitigating the distribution gap between the real and PT3D datasets is essential for various reasons. The primary reason is the generalization of our results. We cannot claim that our results can generalize to one domain if we operate on a completely different domain. Hence, we attempt to minimize the distribution gap. We try to achieve this goal by optimizing specific parameters in the rendering pipeline, as described below.

\subsection{Gaussian Blur}
\begin{figure}
    \centering
    \includegraphics[width=\columnwidth]{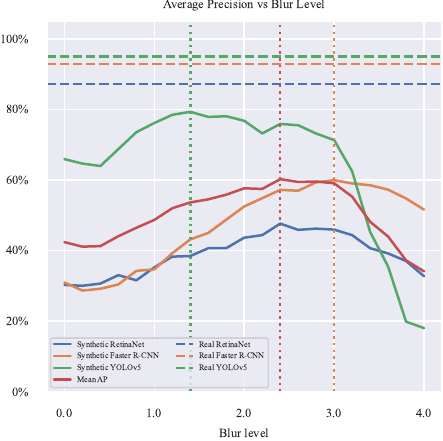}
    \caption{Each point on the solid lines corresponds to a synthetic model trained using the corresponding blur level. Each such model is evaluated on the real validation set. The horizontal dashed lines represent the real models' performance on the real validation set. The vertical dotted lines represent the maxima. The red line represents the average curve of the other three curves. As shown by this analysis, $\sigma=2.4$ is the optimal blur level.}
    \label{fig:blur_analysis}
\end{figure}

We apply blurring to the vehicles in the images to simulate the blurring in the real images. Given a background image $I_\text{bg}$ and the corresponding foreground image (\ie, containing vehicles) $I_\text{fg}$, we apply Gaussian blur:
% \stan{G should not be written here, because you are explaining it after the equation}
\begin{align}
    I_\text{blur} = I_\text{bg} + G(I_\text{fg} - I_\text{bg}),
\end{align}
where $G(\cdot)$ is a Gaussian blur operator with kernel size defined by $k=6\cdot\ceil[\big]{\sigma}-1$, where $\ceil[\big]{\cdot}$ is the ceiling operator, and $\sigma$ is the blur level. We find that $\sigma=2.4$ is the optimal blurring value as shown in \Cref{fig:blur_analysis}. Our analysis of the blur level shows that deficient levels of blur (\ie, close to coarse PT3D renderings) result in less robust synthetic models when evaluated on the real data. Similarly, very high levels of blur (\ie, almost vanished vehicles) also result in poor performance. As expected, the optimal value is somewhere in the middle.
See the effect of applying blurring in \Cref{fig:anti-aliasing_blurring}.

\subsection{Anti-aliasing}
\begin{figure*}
    \centering
    \includegraphics[width=\textwidth]{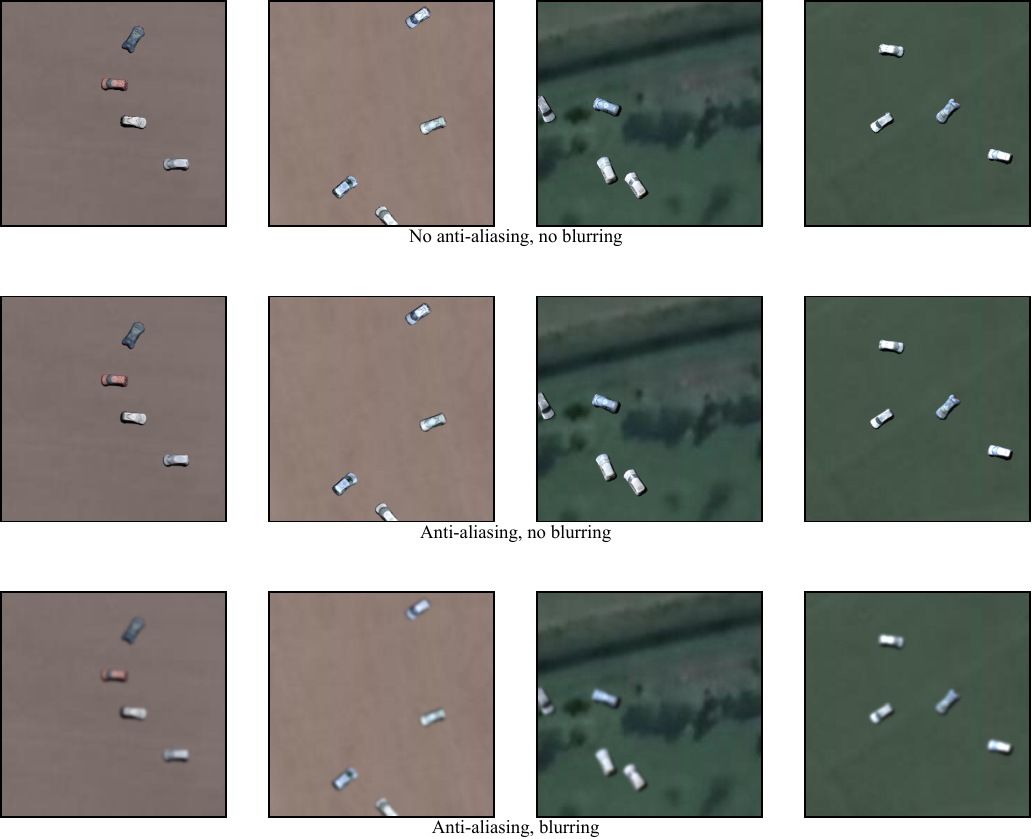}
    \caption{The first row represents the coarse renderings by PyTorch3D. The second row represents the result of applying anti-aliasing. The third row represents the result of applying both anti-aliasing and blurring.}
    \label{fig:anti-aliasing_blurring}
\end{figure*}

We use anti-aliasing techniques to remove pixelization from PyTorch3D's coarse renderings. We apply them by rendering images four times larger than the intended size, then compressing them with the average pooling operator with kernel size $4$ and stride $4$.
See the effect of anti-aliasing in \Cref{fig:anti-aliasing_blurring}.

% \begin{figure*}
%     \centering
%     \includegraphics[width=\textwidth]{Figures/AABlurring.pdf}
%     \caption{The first row represents the coarse renderings by PyTorch3D. The second row represents the result of applying anti-aliasing. The third row represents the result of applying both anti-aliasing and blurring.}
%     \label{fig:anti-aliasing_blurring}
% \end{figure*}

% \begin{figure}
%     \centering
%     \includegraphics[width=\columnwidth]{Figures/Blur-Analysis.pdf}
%     \caption{Each point on the solid lines corresponds to a synthetic model trained using the corresponding blur level. Each such model is evaluated on the real validation set. The horizontal dashed lines represent the real models' performance on the real validation set. The vertical dotted lines represent the maxima. The red line represents the average curve of the other three curves. As shown by this analysis, $\sigma=2.4$ is the optimal blur level.}
%     \label{fig:blur_analysis}
% \end{figure}
\section{Datasets Information}
\label{sec:supp_DatasetsInformation}

This section provides technical details for the datasets we have sampled/annotated (real) or generated (synthetic) for our experiments.

\subsection{Real Datasets}
We produced two real overhead-view datasets for our project. The images were sampled from two online sources: \textit{Land Information New Zealand}\footnote{\url{https://data.linz.govt.nz/}} (LINZ) and \textit{Google Maps} (GMaps). Since both provide georeferenced imagery, the two image sets were sampled from the exact location in New Zealand - Selwyn\footnote{\url{https://data.linz.govt.nz/layer/51926-selwyn-0125m-urban-aerial-photos-2012-2013/}}.

%%%%% LINZ %%%%%
\subsubsection{LINZ Dataset}

\begin{figure}
    \centering
    \includegraphics[width=\columnwidth]{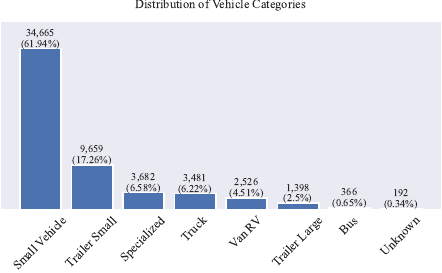}
    \caption{Visualization of vehicle category distribution in the LINZ dataset: each bar signifies the number of samples associated with a specific vehicle category. The figures within the brackets indicate the proportion of total vehicles represented by each class.}
    \label{fig:linz_distribution_classes}
\end{figure}

\begin{figure*}
    \centering
    \includegraphics[width=\textwidth]{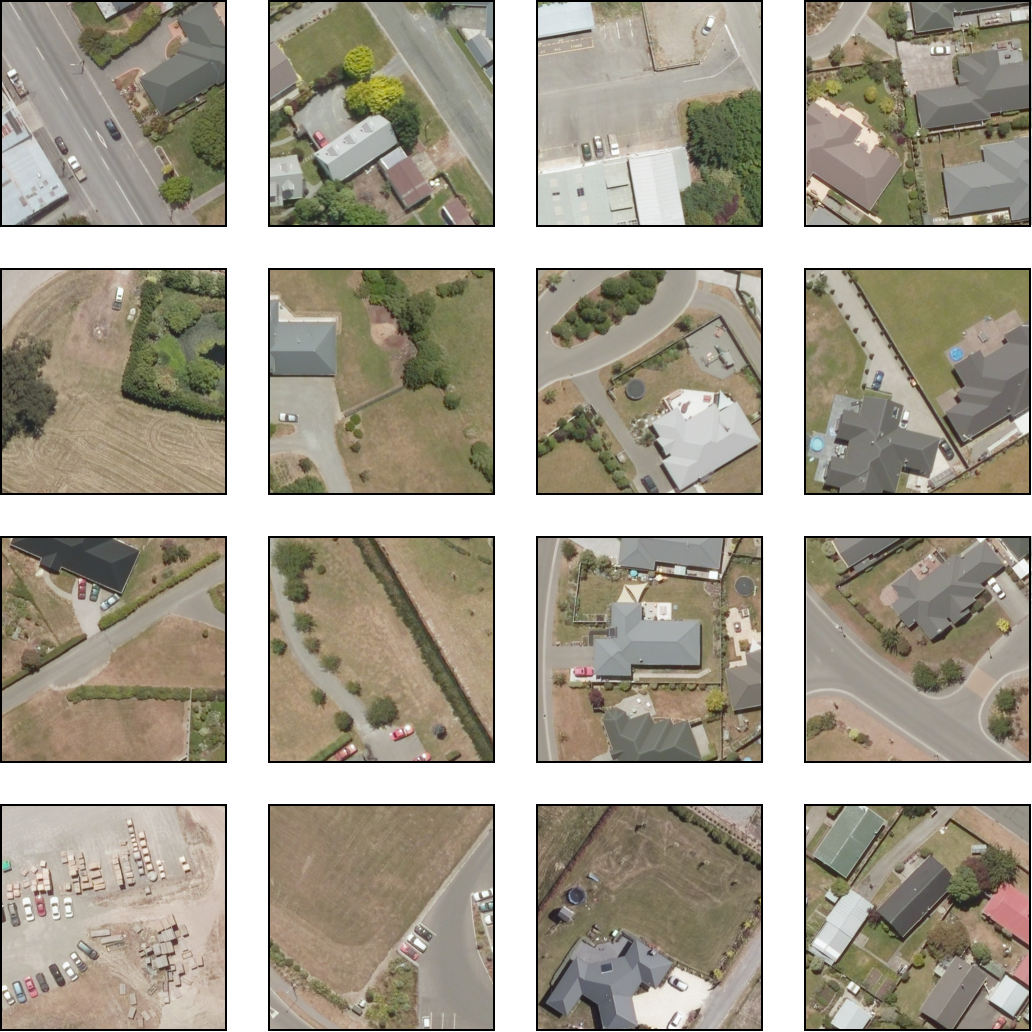}
    \caption{Examples of the labeled LINZ dataset.}
    \label{fig:linz_labeled}
\end{figure*}

\begin{figure*}
    \centering
    \includegraphics[width=\textwidth]{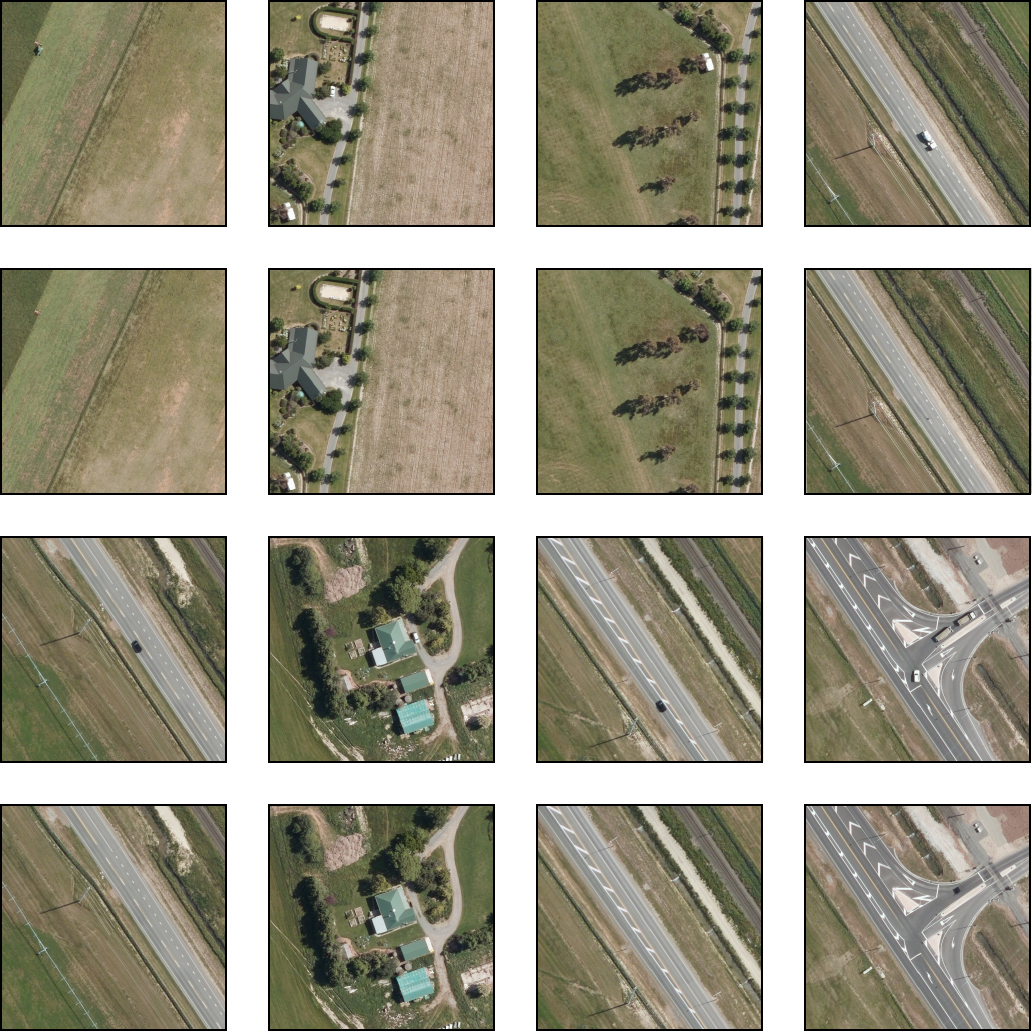}
    \caption{The odd rows represent original images from the LINZ dataset. The even rows represent the corresponding background LINZ images, where the vehicles have been automatically removed.}
    \label{fig:linz_background}
\end{figure*}

Examples of the labeled LINZ and the background LINZ datasets are shown in \Cref{fig:linz_labeled} and \Cref{fig:linz_background}, respectively. 
The distribution between negative (\ie, empty) and positive (\ie, non-empty) images in the LINZ dataset is as follows: \num[]{158944} for negative images and \num[]{13651} for positive images.
See the distribution of vehicle categories in this set of images in \Cref{fig:linz_distribution_classes}. 

% \begin{figure*}
%     \centering
%     \includegraphics[width=\textwidth]{Figures/LINZ-with-vehicles-examples.pdf}
%     \caption{Examples of the labeled LINZ dataset.}
%     \label{fig:linz_labeled}
% \end{figure*}

% \begin{figure*}
%     \centering
%     \includegraphics[width=\textwidth]{Figures/Background-LINZ-examples.pdf}
%     \caption{The odd rows represent original images from the LINZ dataset. The even rows represent the corresponding background LINZ images, where the vehicles have been automatically removed.}
%     \label{fig:linz_background}
% \end{figure*}

% \begin{figure}
%     \centering
%     \includegraphics[width=\columnwidth]{Figures/VehicleCategoryDistribution.pdf}
%     \caption{Visualization of vehicle category distribution in the LINZ dataset: each bar signifies the number of samples associated with a specific vehicle category. The figures within the brackets indicate the proportion of total vehicles represented by each class.}
%     \label{fig:linz_distribution_classes}
% \end{figure}

%%%%% GMAPS %%%%%
\subsubsection{Google Maps (GMaps) Dataset}

\begin{figure*}
    \centering
    \includegraphics[width=\textwidth]{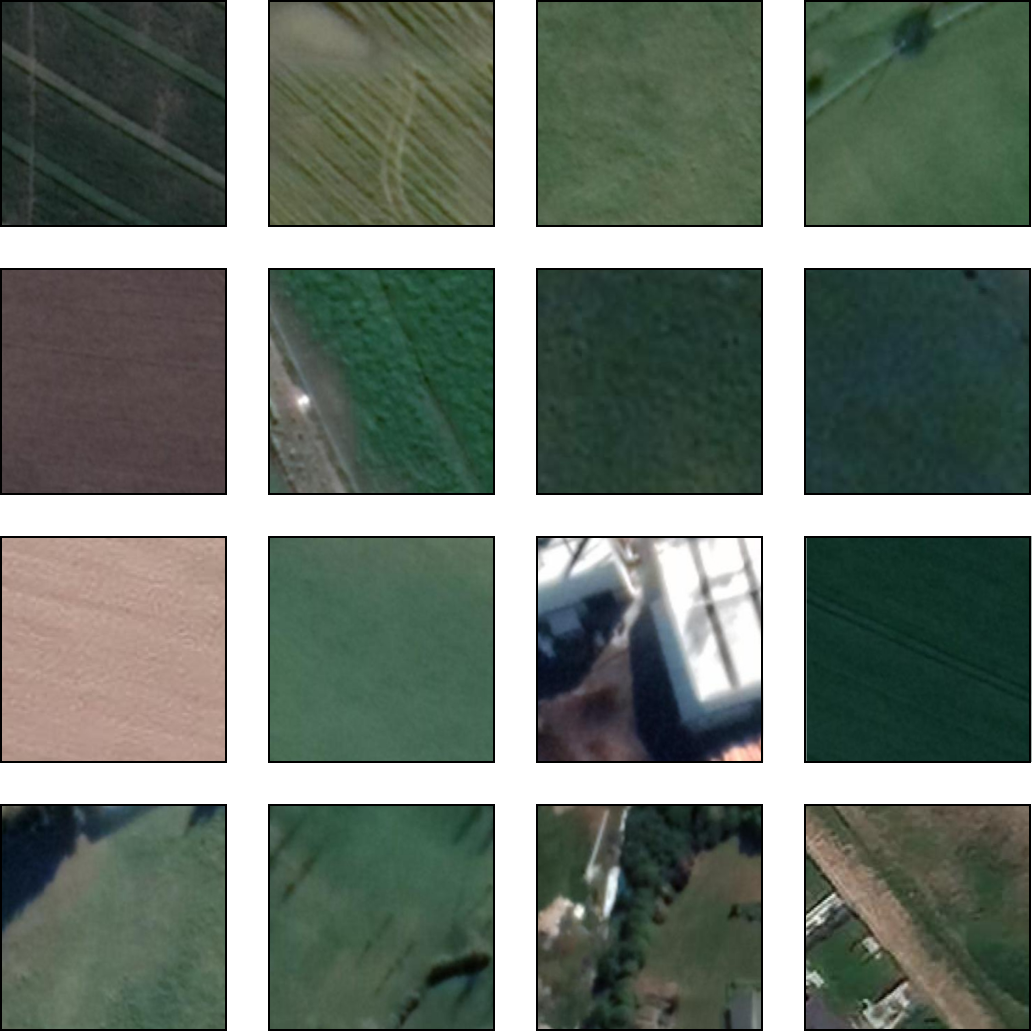}
    \caption{Examples of the GMaps background dataset.}
    \label{fig:gmaps_examples}
\end{figure*}

% We retrieved \num[]{173264} images in total for the GMaps dataset. This number differs from the number of images in the LINZ dataset due to different sampling methods employed in obtaining images for these two datasets. Specifically, when partitioning the larger images sourced from Google Maps, we adopt a random sampling approach for the position and rotation of the cropping window within these larger images. Consequently, this method introduces some overlaps in the final images, resulting in an increased count of the total images.
% See examples of the GMaps dataset images in \Cref{fig:gmaps_examples}.
We retrieved \num[]{173264} images in total for the GMaps dataset, which approximately matches the number of sampled LINZ images.
See examples of the GMaps dataset images in \Cref{fig:gmaps_examples}.

% \begin{figure*}
%     \centering
%     \includegraphics[width=\textwidth]{Figures/GMaps-examples.pdf}
%     \caption{Examples of the GMaps background dataset.}
%     \label{fig:gmaps_examples}
% \end{figure*}

\subsection{Synthetic Datasets}

For our experiments, we rendered various synthetic datasets with original and adversarial objects using two rendering techniques: PyTorch3D and Blender. Here, we provide additional technical details and examples from each.

%%%%% PT3D %%%%%
\subsubsection{PyTorch3D Datasets}

\begin{figure*}
    \centering
    \includegraphics[width=\textwidth]{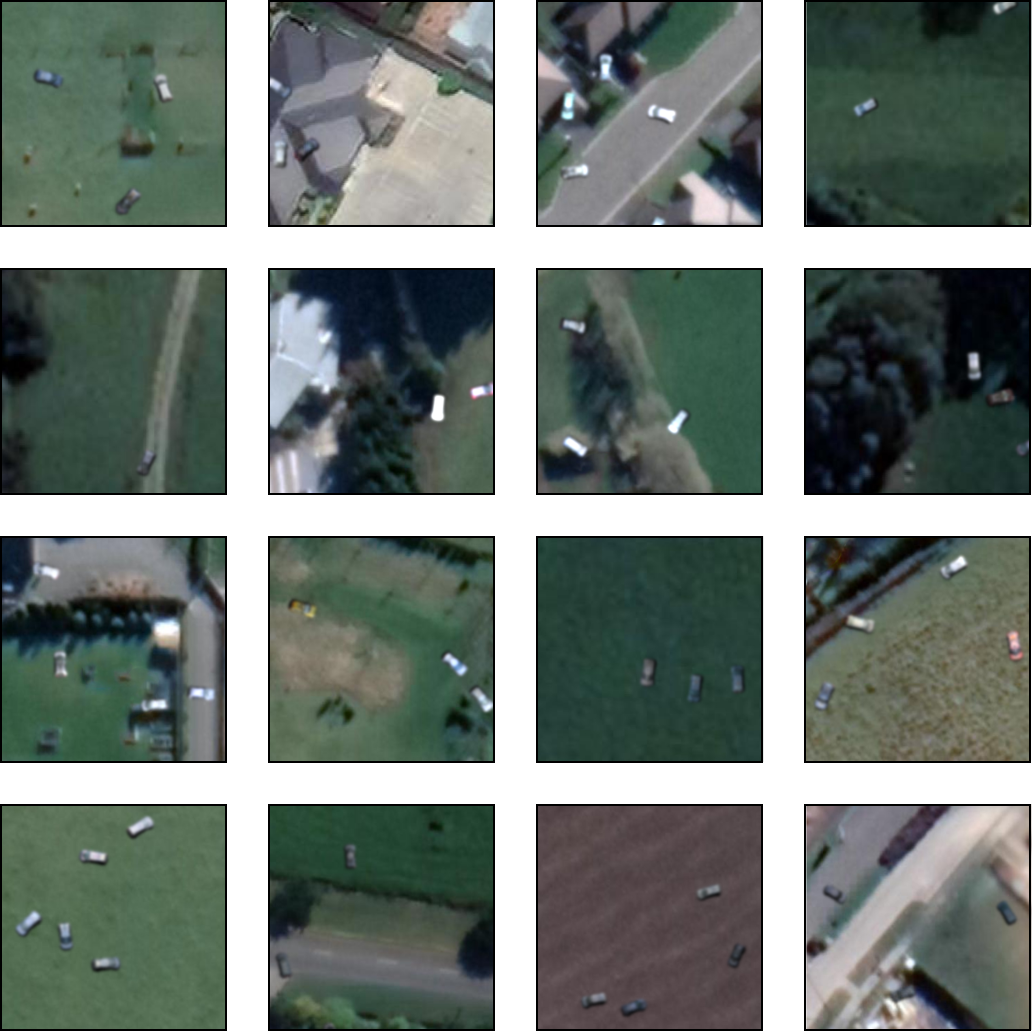}
    \caption{Examples of the original PT3D images.}
    \label{fig:pt3d_original_examples}
\end{figure*}

\textbf{Original.}
This dataset includes original (unmodified) car meshes.
See examples of the PyTorch3D original images in \Cref{fig:pt3d_original_examples}.

% \begin{figure*}
%     \centering
%     \includegraphics[width=\textwidth]{Figures/PT3D-examples.pdf}
%     \caption{Examples of the original PT3D images.}
%     \label{fig:pt3d_original_examples}
% \end{figure*}

\begin{figure*}
    \centering
    \includegraphics[width=\textwidth]{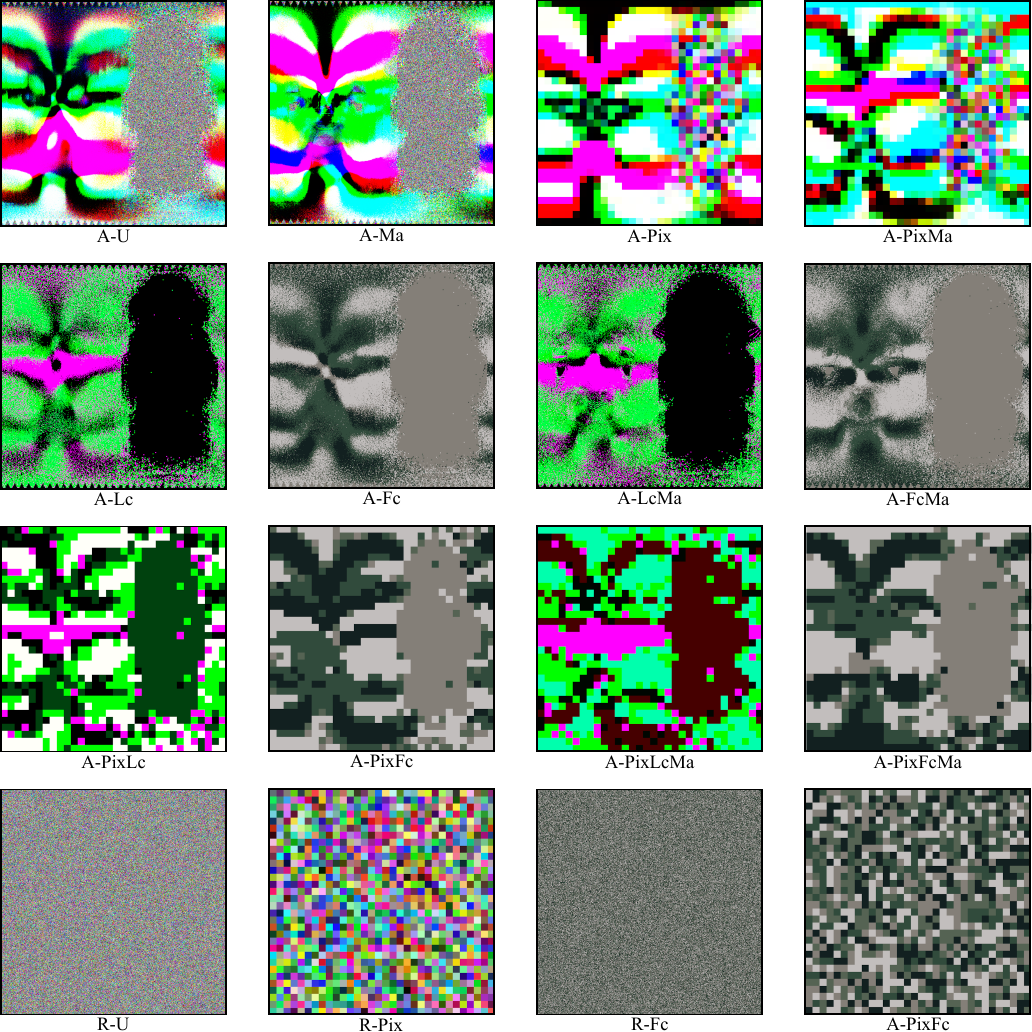}
    \caption{Adversarial and random texture maps. The right side corresponds to the car's underside, which the adversarial optimization ignores.}
    \label{fig:adv_rand_texture_maps}
\end{figure*}

\begin{figure*}
    \centering
    \includegraphics[width=\textwidth]{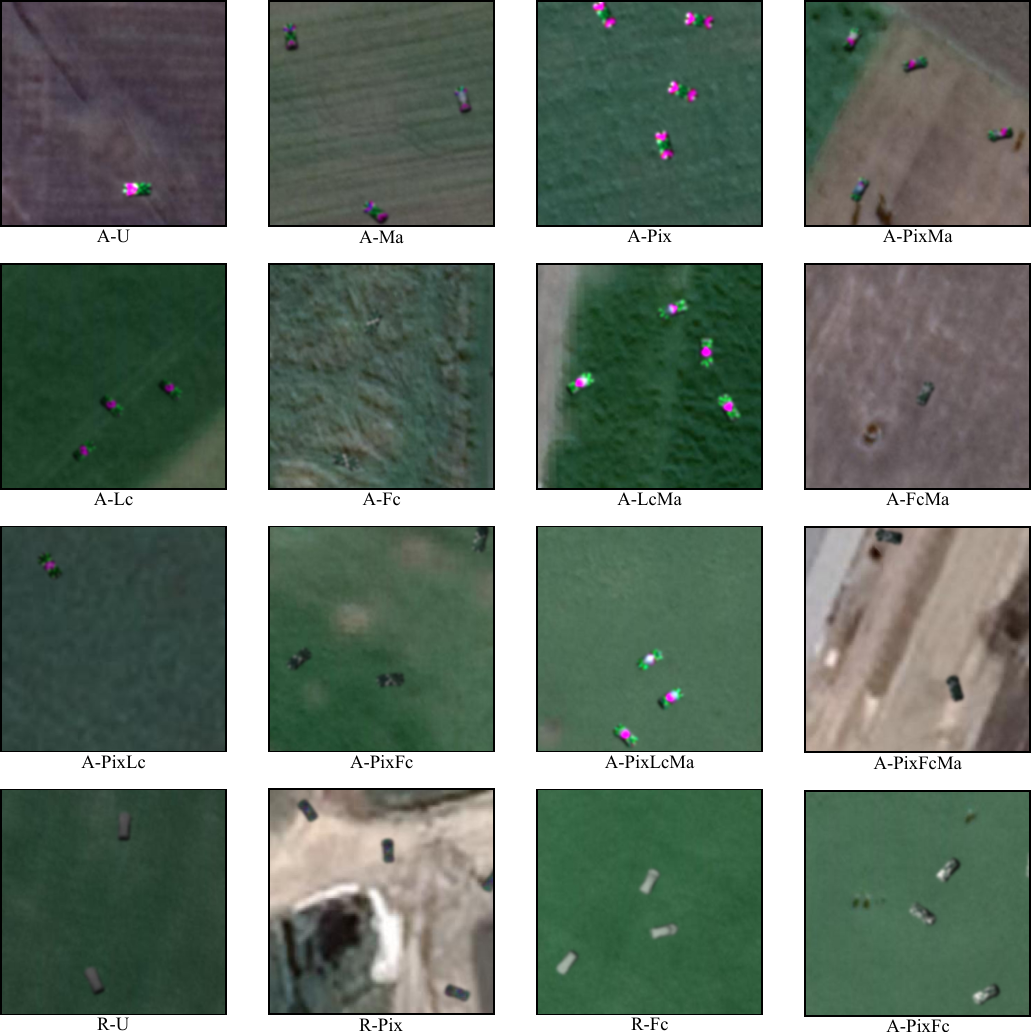}
    \caption{Illustrations of vehicles sourced from the PT3D datasets featuring adversarial and random texture maps.}
    \label{fig:adv_rand_dataset_examples_pt3d}
\end{figure*}

\textbf{Adversarial and Random Textures.}
As described in the paper, we produce twelve adversarial texture maps and four random texture maps. See these texture maps in \Cref{fig:adv_rand_texture_maps}. 
We generate \num[]{5000} validation images for each texture map that we use for evaluation. To generate an image, we first render the meshes using PT3D and then insert a background image sampled from the GMaps dataset. For each scene, we uniformly sample from one to five vehicles. We uniformly randomize the vehicle position and rotation in the scene. The camera always points to the origin of the coordinates as defined in PT3D. To sample the camera pose, we first uniformly sample a 2D-coordinate on a square, which we then re-project on a hemisphere, ensuring that the maximum elevation angle deviation from the vertical position is \ang{20}. 
View examples of these images in \Cref{fig:adv_rand_dataset_examples_pt3d}.

% \begin{figure*}
%     \centering
%     \includegraphics[width=\textwidth]{Figures/AdvRand-Texture-Maps.pdf}
%     \caption{Adversarial and random texture maps. The right side corresponds to the car's underside, which the adversarial optimization ignores.}
%     \label{fig:adv_rand_texture_maps}
% \end{figure*}

% \begin{figure*}
%     \centering
%     \includegraphics[width=\textwidth]{Figures/SupplMat_Textures}
%     \caption{Visualizations of the textures from \Cref{fig:adv_rand_texture_maps} applied to a car mesh.}
%     \label{fig:adv_rand_textures_cars}
% \end{figure*}

% \begin{figure*}
%     \centering
%     \includegraphics[width=\textwidth]{Figures/AdvRand-examples.pdf}
%     \caption{Illustrations of vehicles sourced from the PT3D datasets featuring adversarial and random texture maps.}
%     \label{fig:adv_rand_dataset_examples_pt3d}
% \end{figure*}

%%%%% BLENDER %%%%%
\subsubsection{Blender Datasets}
\label{sec:supp_DatasetsInformation:blender}

\begin{figure*}
    \centering
    \includegraphics[width=\textwidth]{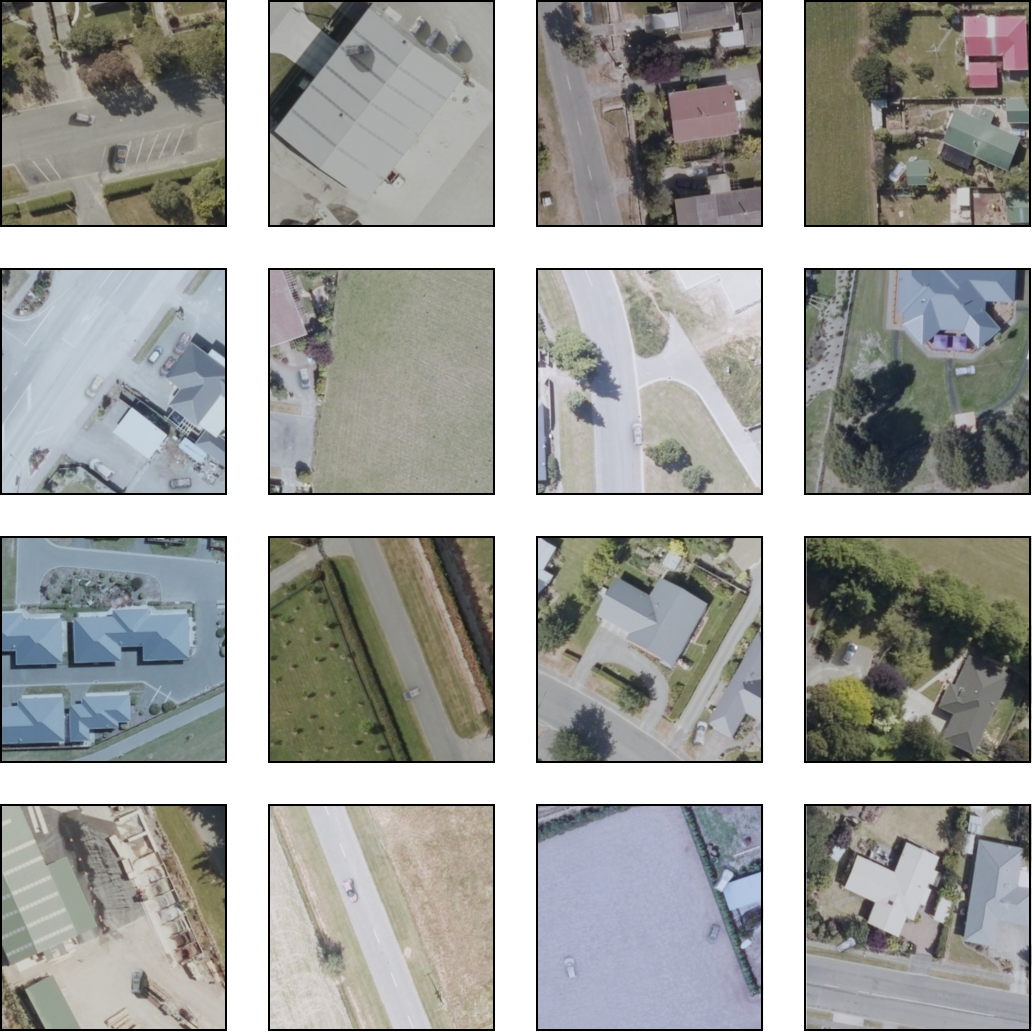}
    \caption{Examples of the original Blender images.}
    \label{fig:blender_original_examples}
\end{figure*}

\textbf{Original.}
We render \num[]{14459} images in Blender, where \num[]{998} contain vehicles and \num[]{13461} images are empty. There are \num[]{2096} vehicles in total in the Blender data.
See examples of the original Blender images in \Cref{fig:blender_original_examples}.

% \begin{figure*}
%     \centering
%     \includegraphics[width=\textwidth]{Figures/Blender-examples.pdf}
%     \caption{Examples of the original Blender images.}
%     \label{fig:blender_original_examples}
% \end{figure*}

\begin{figure*}
    \centering
    \includegraphics[width=\textwidth]{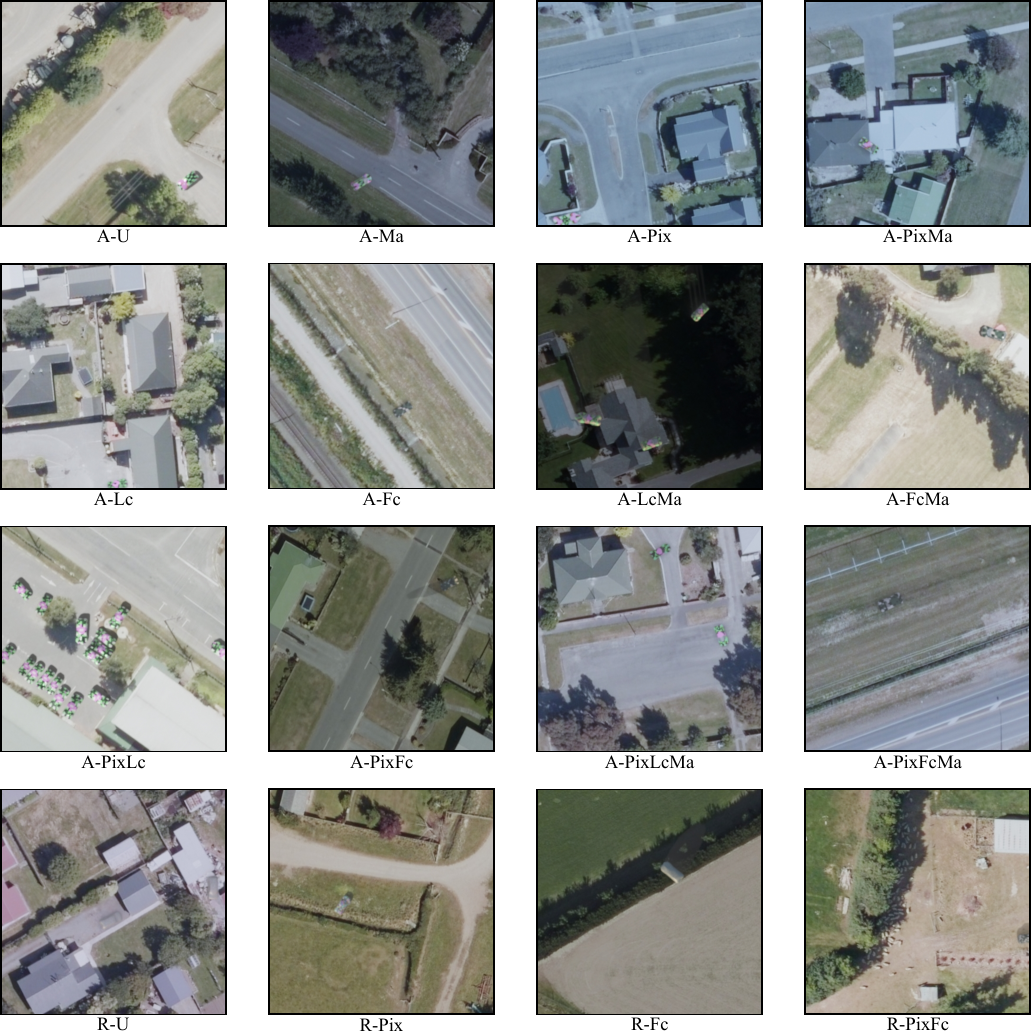}
    \caption{Illustrations of vehicles sourced from the Blender datasets featuring adversarial and random texture maps.}
    \label{fig:adv_rand_dataset_examples_blender}
\end{figure*}

\textbf{Adversarial and Random Textures.}
We use the same adversarial and random texture maps as described for the PT3D adversarial data. We also use the same scenes as for the original Blender data, \ie, \num[]{14459} images, out of which \num[]{998} images contain \num[]{2096} vehicles in total. See example images in \Cref{fig:adv_rand_dataset_examples_blender}.

% \begin{figure*}
%     \centering
%     \includegraphics[width=\textwidth]{Figures/AdvRand-Blender-examples.pdf}
%     \caption{Illustrations of vehicles sourced from the Blender datasets featuring adversarial and random texture maps.}
%     \label{fig:adv_rand_dataset_examples_blender}
% \end{figure*}

% \begin{figure*}
%     \centering
%     \includegraphics[width=\textwidth]{Figures/SupplMat_MeshAttacks.pdf}
%     \caption{Visualization of different shape-based attacks and their corresponding displacement maps.}
%     \label{fig:shape_attacks_vis}
% \end{figure*}
\section{3D Mesh-based Adversarial Attacks}
\label{sec:supp_3DMeshAAs}

In this section, we describe the technical aspects detailing the methodology employed for executing adversarial attacks within each specific setting.

\subsection{Ensemble Attacks}

See \Cref{tab:ensemble_details} for an overview of some significant hyper-parameters related to each adversarial attack reported in the paper.

\begin{table}[h]
    \footnotesize
    \centering
    \caption{An overview of the hyper-parameters used in the ensemble attacks. The loss coefficients $\lambda_1$, $\lambda_2$ and $\lambda_3$ correspond to the loss coefficients applied to the loss objectives of RetinaNet, Faster R-CNN and YOLOv5 respectively, as described in Equation 1 in the main paper.}
    \begin{tabularx}{\columnwidth}
    {
    >{\raggedright\arraybackslash}m{2cm} |
    >{\centering\arraybackslash}X |
    >{\centering\arraybackslash}X |
    >{\centering\arraybackslash}X |
    >{\centering\arraybackslash}X
    }
        \multirow{2}{*}{Attack Type} & \multicolumn{3}{c|}{Loss coefficient $\lambda_i$} & \multirow{2}{*}{\# of epochs} \\
         \cline{2-4}
         & $\lambda_1$ & $\lambda_2$ & $\lambda_3$ & \\
         \hline
         A-U            & 0.020 & 10.000 & 10.000 & 3 \\
         A-Ma           & 0.020 & 10.000 & 10.000 & 3 \\
         A-Pix          & 0.020 & 10.000 & 10.000 & 3 \\
         A-PixMa        & 0.020 & 10.000 & 10.000 & 3 \\
         A-Lc           & 0.020 & 10.000 & 10.000 & 2 \\
         A-Fc           & 0.002 & 10.000 & 2.500  & 2 \\
         A-LcMa         & 0.007 & 10.000 & 5.000  & 2 \\
         A-FcMa         & 0.002 & 10.000 & 2.000  & 2 \\
         A-PixLc        & 0.020 & 10.000 & 10.000 & 2 \\
         A-PixFc        & 0.002 & 10.000 & 2.500  & 2 \\
         A-PixLcMa      & 0.020 & 10.000 & 5.000  & 2 \\
         A-PixFcMa      & 0.003 & 10.000 & 2.000  & 2 \\
         Shape Attack   & 0.020 & 15.000 & 32.000 & 2 \\
         A-Fc (seq.)    & 0.020 & 10.000 & 30.000 & 2 \\
         A-PixFc (seq.) & 0.013 & 18.740 & 20.444 & 2 \\
         A-Fc (par.)    & 0.020 & 10.000 & 30.000 & 2 \\
         A-PixFc (par.) & 0.011 & 13.180 & 20.860 & 2 \\
    \end{tabularx}
    \label{tab:ensemble_details}
\end{table}

\subsection{Texture Optimization}

As outlined in Section 5.2, we employ three constraints that lead to the twelve adversarial texture settings discussed in the main paper. These constraints are \textit{Spatial Resolution}, \textit{Spatial Restriction} and \textit{Color Restriction}. In this section, we discuss the implementation of each constraint.
Before delving into the details of each constraint, we first describe how the adversarial texture is defined in the unconstrained attack (T-U). To execute this attack, a tensor of dimensions $512\times512\times3$ is initialized, representing the adversarial texture map. During this initialization process, each element in the tensor is uniformly sampled from 0 to 1. During the unconstrained attack, this tensor is the optimized entity.

\subsubsection{Spatial Resolution}

To implement the spatial resolution constraint, we store the adversarial texture as a tensor of a smaller size. In our case, because we apply pixelization of size \qtyproduct{16x16}{px}, we store a latent representation of the adversarial texture as a $32\times32\times3$ tensor, where the first two dimensions are derived from the fact that the final texture map is expected to be $512\times512\times3$, hence $512 / 16 = 32$. Upon texture generation request, we upscale this tensor to $512\times512\times3$ using the nearest-neighbor interpolation, resulting in a pixelated output.

\subsubsection{Spatial Restriction}

\begin{figure*}
    \centering
    \includegraphics[width=\textwidth]{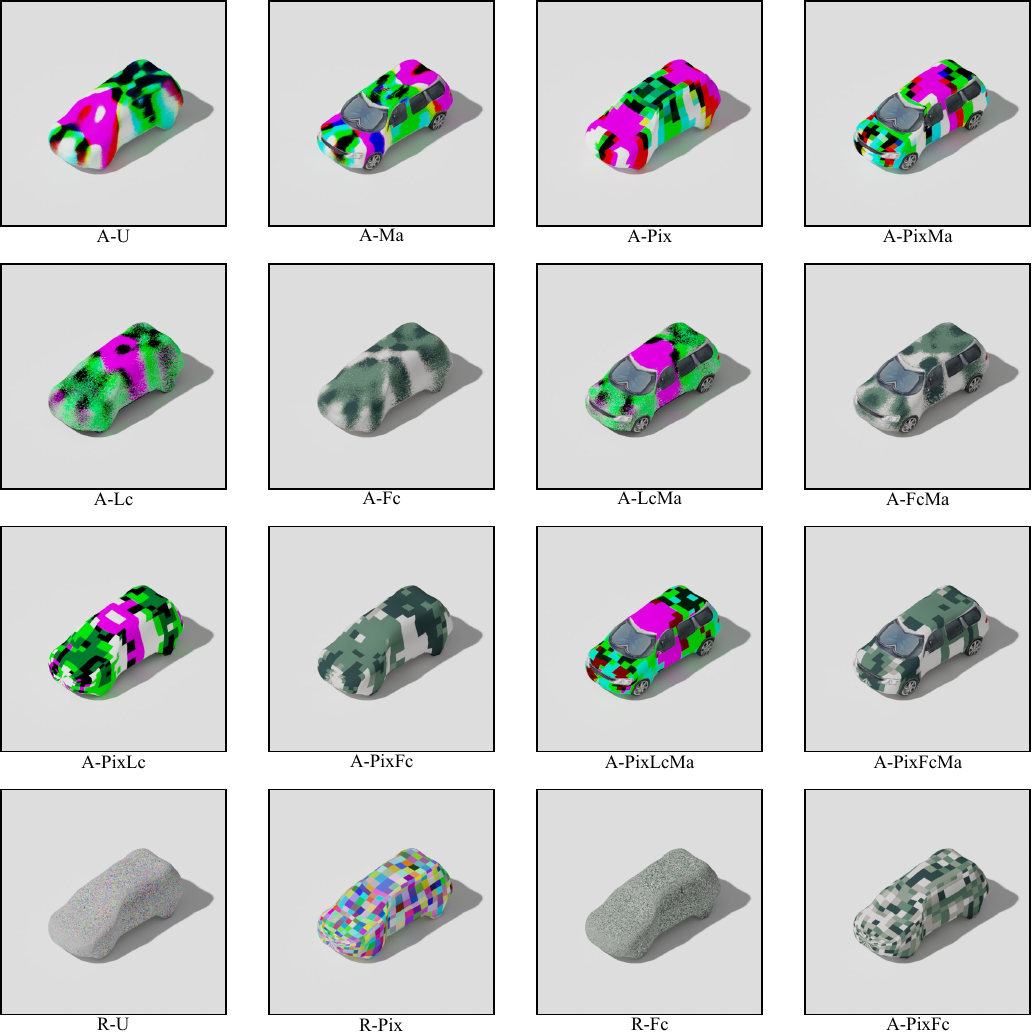}
    \caption{Visualizations of the textures from \Cref{fig:adv_rand_texture_maps} applied to a car mesh.}
    \label{fig:adv_rand_textures_cars}
\end{figure*}

To implement the spatial restriction constraint, we use an adversarial texture map $T_\text{adv}$ of size $512\times512\times3$, an original texture map $T_\text{or}$ of a vehicle to which the adversarial texture is applied, and its corresponding binary segmentation mask $T_\text{mask}$. Using these three entities, we produce the segmented adversarial texture map
\begin{align}
    T_\text{segmented} = T_\text{or}\cdot\left(1 - T_\text{mask}\right) + T_\text{adv}\cdot T_\text{mask}.
\end{align}
See the ``Ma'' texture maps in \Cref{fig:adv_rand_textures_cars} to understand the final result. When combining this constraint with the Spatial Resolution constraint, we first produce a pixelated adversarial texture map and then apply masking.

\subsubsection{Color Restriction}

Consider the following example to understand how the color constraint is implemented differently. Let $p_i$ be the $i$-th pixel of a texture map, such that $p_i\in P$, where $P\in\mathbb{R}^{(H_t\cdot W_t)\times 3}$ is the set of all pixels in the texture map of size $\left(H_t\times W_t\times 3\right)$. In addition, let $C=\{c_i,\forall i=1,2,\dots,N\}$, $c_i\in\mathbb{R}^3$ be the limited set of colors that we want to enforce for painting the texture map, where $N$ is the number of allowed colors. 

Ideally, we would like to be able to perform ${\arg\min}$ in a differentiable manner to reassign each pixel value at each attack iteration, such that $p_i \leftarrow \underset{c_i\in C}{\arg\min}(\|p_i - c_i\|_2)$.
However, it is unclear how to do this differently. Therefore, we modify the pipeline to perform it in a differentiable fashion. 
First of all, we change the definition of each pixel in the texture map: instead of representing RGB values, each pixel now represents a set of probabilities of belonging to a particular color $c_i$ from the set of colors $C$, \ie,
$p_i\in \mathbb{R}^{N}$,
$\sum_k p_{i,k}=1$,
$P\in\mathbb{R}^{(H_t\cdot W_t)\times N}$ and
$p_i=(p_{i,1},p_{i,2}\dots,p_{i,N})$, 
where $p_{i,k}$ is the probability that the $i$-th pixel in the texture map is $c_k$.
Second, we define a softmax-like function, which we use to amplify the maximum value in a vector and suppress the non-maximum values. We control the amplification and suppression levels with a temperature parameter $\tau$. Applying this softmax-like function to some vector $r=[r_1,r_2,\dots]^\text{T}$, we obtain $s(r_i) = \frac{e^{\ln{(r_i)}/\tau}}{\sum_j e^{\ln{(r_j)}/\tau}}$.
For simplicity, let $s(p_i)$ represent $[s(p_{i,1}),\dots,s(p_{i,N})]^\text{T}$. Whenever prompted to generate a texture map with the Color Restriction constraint, we perform the following procedure on each pixel $p_i$ to obtain its output RGB form $\hat{p}_i\in\mathbb{R}^3$:
\begin{align}
    \label{eq:class-probability-weights}
    \mathbf{w}_i &= s\left(s\left(p_i\right)\right), \\
    \label{eq:weighted-sum-colors}
    \hat{p}_i &= \mathbf{C}\cdot \mathbf{w}_i,
\end{align}
where $\mathbf{C} = \left[c_1\:c_2\:\dots\:c_N\right]\in\mathbb{R}^{3\times N}$. In other words, we first shift the probabilities towards the maximum probability class as shown in \cref{eq:class-probability-weights}, then, treating probabilities as weights, we perform a weighted sum of colors $C$ as shown in \cref{eq:weighted-sum-colors}. 
As we empirically find, performing soft-argmax (\cref{eq:class-probability-weights,eq:weighted-sum-colors}) \textit{twice} results in a much better approximation to argmax than if it was performed only once. We tried reducing the temperature parameter $\tau$ and performing soft-argmax only once, but lower temperatures resulted in numerical instability. 
After each attack cycle, softmax is applied to each $p_i$ to ensure $\sum_k p_{i,k}=1$. 
After finishing an adversarial attack with this constraint, a non-differentiable ${\arg\max}$ assigns colors from $C$ to each pixel, ensuring a final texture map with at most $N$ colors.
During the attack, either $\{P\}$ or $\{P,C\}$ can be optimized, contingent upon the applied restrictions (``Fc'' or ``Lc'' respectively).

\subsection{Shape Optimization}

As outlined in Section 5.3 in the main paper, we optimize the displacement map, transforming the pixel values into vertex deformations. We employ two constraints \textit{Symmetry} and \textit{Magnitude}.

We initialize a negative displacement tensor of shape $64\times64\times1$ representing a single channel (grayscale) image. This ensures that the number of pixels in the displacement map (\num[]{4096}) is always greater than the number of vertices of the car meshes we use (\num[]{1000}). Contrary to the texture map initialization, we initialize this tensor with zeros, aiming to start from the un-deformed state. 

Additionally, a topology map is calculated from the static UV map of each mesh. The topology map gathers and retains the information of each unique vertex from the UV map. This helps align the displacement maps, UV maps, and texture maps. 

The deformation at each vertex of the mesh is calculated by using the equation
\begin{align}
\label{eq:deformation-disp-maps}
    \Delta V_\text{i} = R_i\cdot D_i,
\end{align}
where $R_i$ is the vector defined by joining the geometric mean of all the vertices of the vehicle mesh and the corresponding vertex coordinate $V_i$. To calculate $D_i$, the displacement tensor is circularly padded to match the dimensions of the UV map. Then, each point sampled from the topology map is used to interpolate the aligned displacement map bi-linearly to calculate the corresponding deformation $D_i$. The two constraints are imposed as follows. 

\subsubsection{Symmetry}

\begin{figure}
    \centering
    \includegraphics{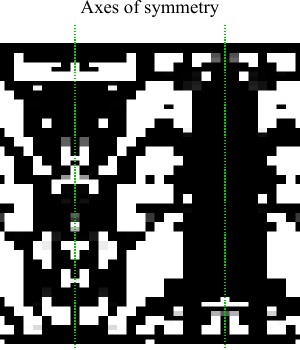}
    \caption{An example of the axes of symmetry in a displacement map. The highlighted axes correspond to the central plane that cuts the mesh in two halves along its longitudinal direction.}
    \label{fig:disp_map_symmetry}
\end{figure}

To ensure the symmetrical layout of the mesh, we apply a symmetry mask while deforming the mesh. This symmetry mask is calculated from the UV map with two axes of symmetry. The axes of symmetry for the corresponding displacement map are shown in \Cref{fig:disp_map_symmetry}.

\subsubsection{Magnitude}

\begin{figure*}
    \centering
    \includegraphics[width=\textwidth]{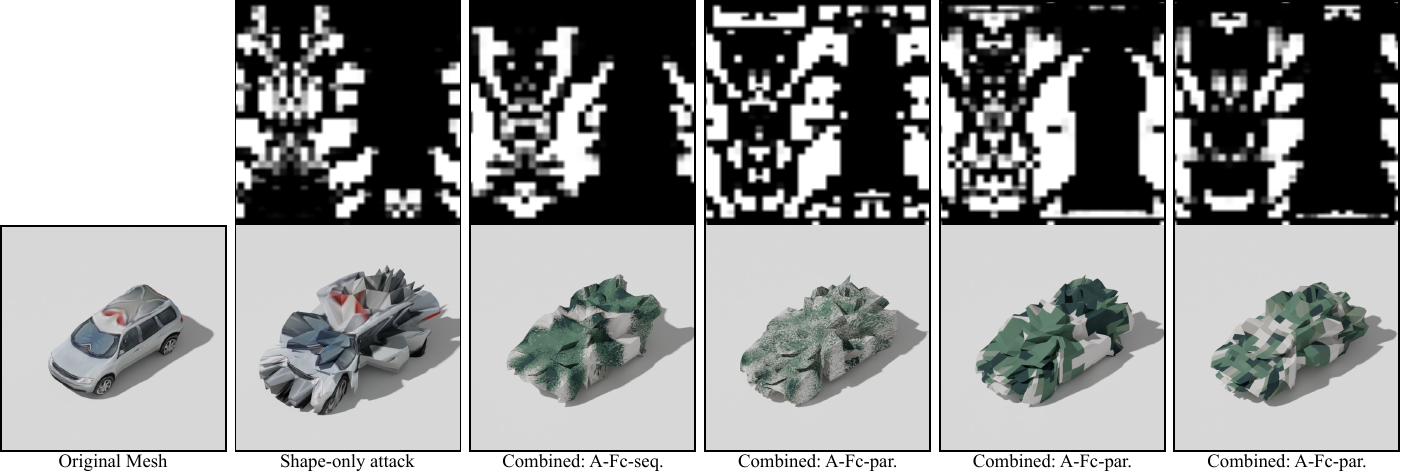}
    \caption{Visualization of different shape-based attacks and their corresponding displacement maps.}
    \label{fig:shape_attacks_vis}
\end{figure*}

The maximum amount of perturbation is crucial to determine the practicality. We determine the width of the car mesh by finding the maximum difference of vertex positions along the width. The sigmoid function $\sigma(x)=1/\left(1+e^{-x}\right)$ is used on the displacement tensor to make sure its values lie between 0 and 1. The final deformations are calculated by extending \Cref{eq:deformation-disp-maps}.
\begin{align}
\label{eq:constrained-deformation-disp-maps}
    \Delta V_i = \text{PM}\cdot W\cdot \sigma\left(R_i\cdot D_i\right)
\end{align}
where $\text{PM}$ is the perturbation magnitude as defined in Section 5.3 in the main paper and $W$ is the width of the car. The displacement maps corresponding to optimal perturbations, when greater than 0, as described in Table 2 in the main paper are shown in \Cref{fig:shape_attacks_vis}. 
\section{Additional Results}
\label{sec:supp_AdditionalResults}

In this section, we report some further results to complement the arguments from the main paper.

\subsection{EASR}
\label{sec:supp_AdditionalResults:EASR}

To evaluate the effectiveness of the adversarial meshes, we rendered two matched image datasets, $D_\text{or}$ and $D_\text{adv}$, from each experiment.  
For $D_\text{or}$, we composed and rendered 3D scenes with the original car meshes. For $D_\text{adv}$, we apply adversarial modifications to the meshes of the same scenes. Thus, each image from $D_\text{or}$ has an identical version (the same background, lighting, camera parameters, and car locations and orientations) with adversarial cars.
We compute the percentage of vehicles \textbf{\underline{d}}etected in $D_\text{or}$ but \textbf{\underline{m}}issed in $D_\text{adv}$. $V_{d,m}$ represents such vehicles, and $V_{d,d}$ denotes vehicles detected in both $D_\text{or}$ and $D_\text{adv}$. This computation yields the \textit{Attack Success Rate} $\text{ASR} = \frac{|V_{d,m}|}{|V_{d,d} \cup V_{d,m}|}$, where $|\cdot|$ is the cardinality operator. In our task, avoiding introducing new detections $V_{m,d}$ after applying the adversarial entity is also important. Thus, we modify ASR to account for this:
\begin{align}
    \text{EASR} &= 
    \label{eq:effective-success-rate}
    \frac{|V_{d,m}| - |V_{m,d}|}{|V_{d,d} \cup V_{d,m}|} =
    \text{ASR} - \text{ER},
\end{align}
where EASR is the \textit{Effective Attack success Rate} and $\text{ER}$ is the \textit{erroneous rate}, \ie fraction of true-positive detections that emerged after introducing the adversarial entity.

\subsection{APD}

In addition to computing the EASR, we evaluate the average precision drop (APD) when running adversarial attacks. 
To calculate APD, we first compute the AP on a dataset of original images $D_\text{or}$ and on a dataset of adversarial images ($D_\text{adv}$ (where both are as defined in Section 7.1 in the main paper), resulting in $\text{AP}_\text{or}$ and $\text{AP}_\text{adv}$ respectively. We then obtain the average precision drop as $\text{APD}=\text{AP}_\text{or}-\text{AP}_\text{adv}$.
See the results in \Cref{tab:texture_results_PT3D_SM}.

As anticipated and previously noted, the findings indicate that introducing constraints diminishes performance while incorporating shape modifications alongside texture alterations restores performance. We also highlight the notably low APD observed in randomly generated texture maps, implying that replicating adversarial modifications randomly may yield poor results.

\begin{table}[h]
        \centering
        \footnotesize
        \caption{
            The figures show mean values from evaluations of individual synthetic models on PT3D and Blender data. ``T'', ``R'', ``S'' and ``C'' represent the texture, random texture, shape, and combined attacks. Note that Lc and Fc are mutually exclusive by definition. The constraints follow the definitions outlined in Section 5.2. $\text{PM}^\star$ and $\text{Pr}^\star$ represent the optimal perturbation magnitude and practicality of the attacks involving shape modifications.
        }
        \label{tab:texture_results_PT3D_SM}
        \begin{tabularx}{\columnwidth}{
            >{\raggedright\arraybackslash}m{17mm}| 
            >{\centering\arraybackslash}m{4mm}| 
            >{\centering\arraybackslash}m{4mm}| 
            >{\centering\arraybackslash}m{4mm}| 
            >{\centering\arraybackslash}m{4mm}| 
            >{\centering\arraybackslash}m{5mm}| 
            >{\centering\arraybackslash}m{5mm}|
            >{\centering\arraybackslash}X| 
            >{\centering\arraybackslash}X
        }
    
            \multirow{2}{*}{\parbox{17mm}{\centering\textbf{Attack}}} & 
            \multicolumn{4}{c|}{\textbf{Constraints}} & 
            $\text{PM}^\star$ &
            $\text{Pr}^\star$ &
            \textbf{PT3D} &
            \textbf{Blender} \\
            \cline{2-5}
            
             & Pix & Lc & Fc & Ma & & &
             \textbf{APD} & 
             \textbf{APD} \\
            \hline
            T-U & & & &                                         & --- & --- & 50.97\% & 63.17\% \\
            T-Ma & & & & \checkmark                             & --- & --- & 32.68\% & 40.67\% \\
            T-Pix & \checkmark & & &                            & --- & --- & 53.28\% & 59.03\% \\
            T-PixMa & \checkmark & & & \checkmark               & --- & --- & 24.20\% & 37.47\% \\
            T-Lc & & \checkmark & &                             & --- & --- & 46.17\% & 64.95\% \\
            T-Fc & & & \checkmark &                             & --- & --- & 7.54\% & 48.46\% \\
            T-LcMa & & \checkmark & & \checkmark                & --- & --- & 26.83\% & 46.80\% \\
            T-FcMa & & & \checkmark & \checkmark                & --- & --- & 2.56\% & 23.11\% \\
            T-PixLc & \checkmark & \checkmark & &               & --- & --- & 56.30\% & 62.13\% \\
            T-PixFc & \checkmark & & \checkmark &               & --- & --- & 9.62\% & 48.92\% \\
            T-PixLcMa & \checkmark & \checkmark & & \checkmark  & --- & --- & 22.97\% & 37.59\% \\
            T-PixFcMa & \checkmark & & \checkmark & \checkmark  & --- & --- & 2.53\% & 38.32\% \\
            \hline
            R-U & & & &                                         & --- & --- & 0.21\% & 13.69\% \\
            R-Pix & \checkmark & & &                            & --- & --- & 0.51\% & 16.77\% \\
            R-Fc & & & \checkmark &                             & --- & --- & 0.85\% & 15.82\% \\
            R-PixFc & \checkmark & & \checkmark &               & --- & --- & 0.62\% & 17.75\% \\
            \hline
            S-O & --- & --- & --- & ---                         & 0.4 & 0.6 & 53.18\% & 72.47\% \\
            \hline
            C-U & & & &                                         & 0.0 & 1.0 & 55.43\% & --- \\
            C-Pix & \checkmark & & &                            & 0.0 & 1.0 & 58.09\% & --- \\
            C-Lc & & \checkmark & &                             & 0.0 & 1.0 & 50.49\% & --- \\
            C-Fc (seq.) & & & \checkmark &                      & 0.2 & 0.8 & 18.35\% & 62.96\% \\
            C-Fc (par.) & & & \checkmark &                      & 0.2 & 0.8 & 37.39\% & 62.57\% \\
            C-PixLc & \checkmark & \checkmark & &               & 0.0 & 1.0 & 58.13\% & --- \\
            C-PixFc (seq.) & \checkmark & & \checkmark &        & 0.2 & 0.8 & 36.09\% & 68.64\% \\
            C-PixFc (par.) & \checkmark & & \checkmark &        & 0.2 & 0.8 & 37.79\% & 71.18\% \\
    \end{tabularx}
\end{table}

\subsection{Original Blender Data Evaluation}

We also report the evaluation results of all models using the original Blender data. See example images in \Cref{sec:supp_DatasetsInformation:blender}, and the evaluation results in \Cref{tab:original_blender_results}.
The evaluation results suggest that almost all models perform quite well on the Blender original data. It could be the consequence of the Blender dataset being a high quality simulation of real world data, \ie it is between the coarse PT3D data and the fine-grained LINZ data. Consequently, both the real and synthetic models exhibit strong performance, attributed to the close resemblance between the Blender dataset and the training sets of both sets of models.

\subsection{Evaluating Real Data Models on the Adversarial Data}

\begin{figure*}
    \centering
    \includegraphics[width=\textwidth]{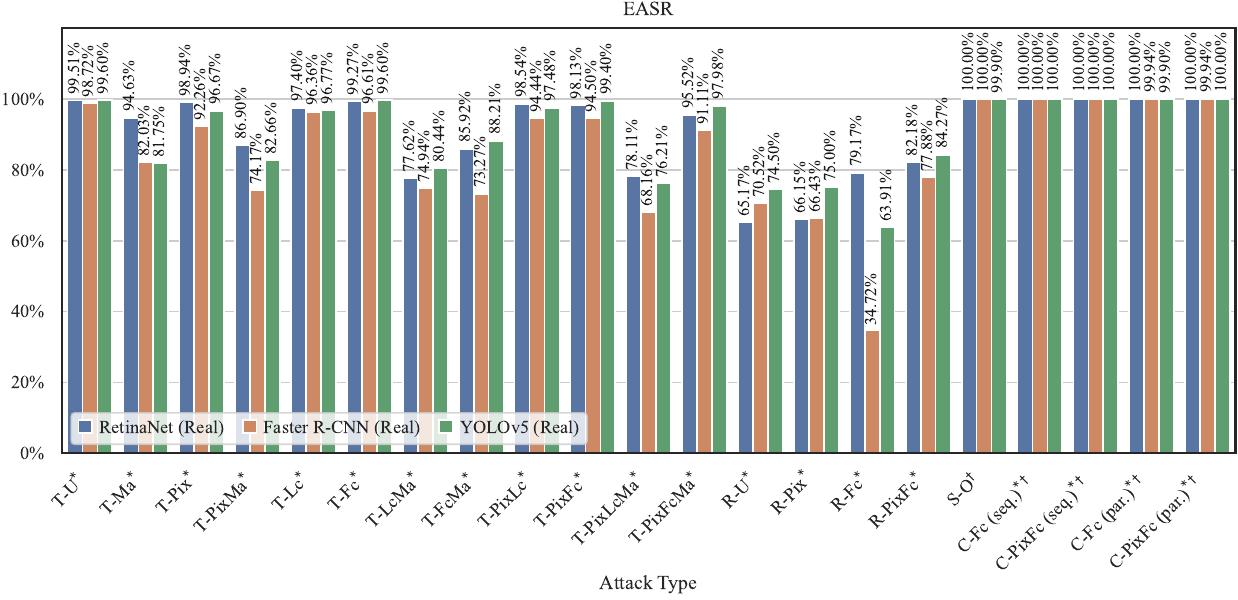}
    \caption{Evaluation results of the models trained on real data and tested on the Blender-generated adversarial datasets.\\Dentations: $^{*}$texture-only attacks, $^{\dagger}$shape-only attacks, and $^{*\dagger}$combined attacks.}
    \label{fig:real_models_results_1}
\end{figure*}

\begin{figure*}
    \centering
    \includegraphics[width=\textwidth]{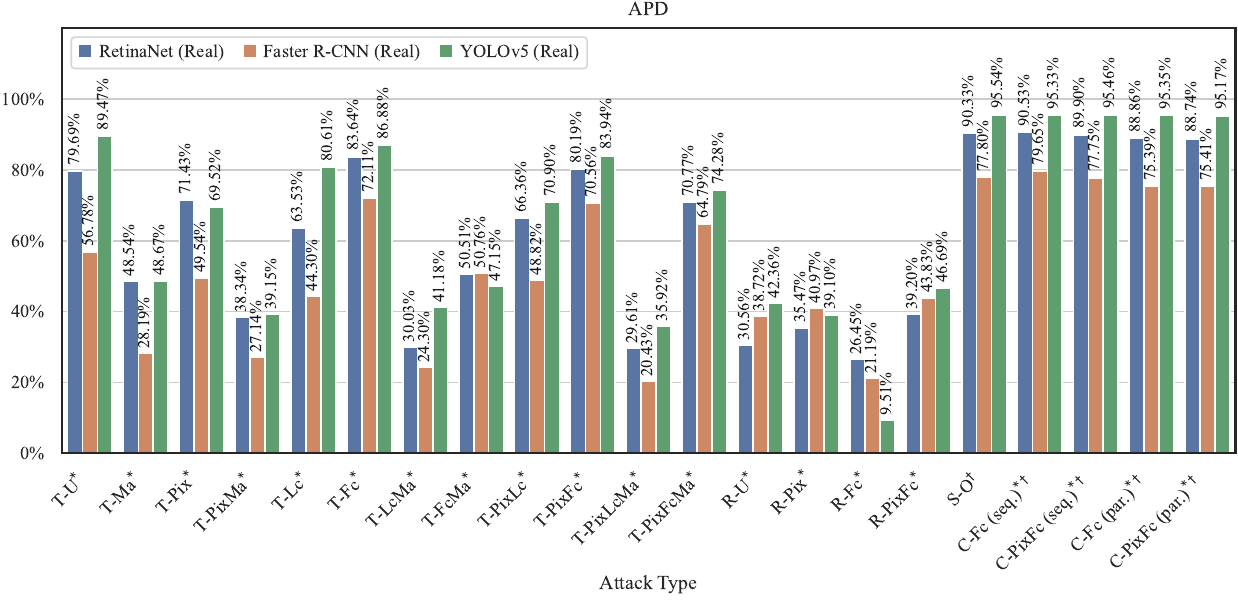}
    \caption{Evaluation results of the models trained on real data and tested on the Blender-generated adversarial datasets.\\Dentations: $^{*}$texture-only attacks, $^{\dagger}$shape-only attacks, and $^{*\dagger}$combined attacks.}
    \label{fig:real_models_results_2}
\end{figure*}

We also evaluate the real models (\ie, trained on real data) on all adversarial datasets rendered using Blender. We run these experiments to assess how robust models trained on real data would react to adversarial samples that are highly realistic. However, we recognize the distribution gap between the real data and the synthetically produced data with Blender. See the results of the evaluations in \Cref{fig:real_models_results_1,fig:real_models_results_2}.

\section{Practicality and Comparisons}
\label{sec:SM:practicality_and_comparisons}

This section extends the arguments discussed in Section 6 in the main paper. See the extended version of Table 1 from the main paper below in \Cref{tab:comparisons_SM}.
Because the optimal $\text{Pr}$ level found for C-U, C-Pix, C-Lc, and C-PixLc is $1.0$, \ie $\text{PM}=0.0$, hence these combined attacks are not considered in \Cref{tab:comparisons_SM}, because they correspond to their texture-based counterparts T-U, T-Pix, T-Lc, and T-PixLc, respectively.

\begin{table}[h]
    \centering
    \footnotesize
    \caption{Comparing the practicality of the attacks explored in our study to previous works. We do not distinguish between sequential and parallel combined attacks as they only impact the process, not the final result's form. The first symbol reflects the texture-related practicality score, the second symbol reflects the shape-related practicality score. This table complements Table 1 from the main paper. Compared to the table in the main paper, the ``Notes'' column is missing because we discuss the score of each camouflage in detail in the text, instead of leaving brief notes in the table.}
    \begin{tabularx}{\columnwidth} { 
      >{\centering\arraybackslash}m{3mm}| 
      >{\raggedright\arraybackslash}m{3cm}| 
      >{\centering\arraybackslash}X| 
      >{\centering\arraybackslash}X| 
      >{\centering\arraybackslash}X| 
      >{\centering\arraybackslash}m{1cm}
      }
     & \textbf{Camouflage} & \textbf{PC} & \textbf{DI} & \textbf{DO} & \textbf{Total Score} \\
     \hline
     
     \multirow{6}{*}{\rotatebox{90}{Other works}} 
     & Du \etal (ON) [22]   & $\textcolor{blue}{+}\textcolor{orange}{0}$ & $\textcolor{blue}{+}\textcolor{orange}{0}$ & $\textcolor{blue}{+}\textcolor{orange}{0}$ & $+3$ \\
     \cline{2-6}
     & Du \etal (OFF) [22]  & $\textcolor{blue}{0}\textcolor{orange}{0}$ & $\textcolor{blue}{+}\textcolor{orange}{0}$ & $\textcolor{blue}{-}\textcolor{orange}{0}$ & $0$  \\
     \cline{2-6}
     & EVD4UAV [71]         & $\textcolor{blue}{+}\textcolor{orange}{0}$ & $\textcolor{blue}{+}\textcolor{orange}{0}$ & $\textcolor{blue}{+}\textcolor{orange}{0}$ & $+3$ \\
     \cline{2-6}
     & FCA [75]             & $\textcolor{blue}{-}\textcolor{orange}{0}$ & $\textcolor{blue}{-}\textcolor{orange}{0}$ & $\textcolor{blue}{+}\textcolor{orange}{0}$ & $-1$ \\
     \cline{2-6} 
     & ACTIVE [73]          & $\textcolor{blue}{-}\textcolor{orange}{0}$ & $\textcolor{blue}{-}\textcolor{orange}{0}$ & $\textcolor{blue}{+}\textcolor{orange}{0}$ & $-1$ \\
     \cline{2-6}
     & DTA [72]             & $\textcolor{blue}{-}\textcolor{orange}{0}$ & $\textcolor{blue}{-}\textcolor{orange}{0}$ & $\textcolor{blue}{+}\textcolor{orange}{0}$ & $-1$ \\
     \hline
     
     \multirow{15}{*}{\rotatebox{90}{Our}} 
     & T-U          & $\textcolor{blue}{-}\textcolor{orange}{0}$ & $\textcolor{blue}{-}\textcolor{orange}{0}$ & $\textcolor{blue}{-}\textcolor{orange}{0}$ & $-3$ \\
     \cline{2-6}
     & T-Ma         & $\textcolor{blue}{-}\textcolor{orange}{0}$ & $\textcolor{blue}{-}\textcolor{orange}{0}$ & $\textcolor{blue}{+}\textcolor{orange}{0}$ & $-1$ \\
     \cline{2-6}
     & T-Pix        & $\textcolor{blue}{-}\textcolor{orange}{0}$ & $\textcolor{blue}{+}\textcolor{orange}{0}$ & $\textcolor{blue}{-}\textcolor{orange}{0}$ & $-1$ \\
     \cline{2-6}
     & T-PixMa      & $\textcolor{blue}{-}\textcolor{orange}{0}$ & $\textcolor{blue}{+}\textcolor{orange}{0}$ & $\textcolor{blue}{+}\textcolor{orange}{0}$ & $+1$ \\
     \cline{2-6}
     & T-Lc         & $\textcolor{blue}{-}\textcolor{orange}{0}$ & $\textcolor{blue}{-}\textcolor{orange}{0}$ & $\textcolor{blue}{-}\textcolor{orange}{0}$ & $-3$ \\
     \cline{2-6}
     & T-Fc         & $\textcolor{blue}{-}\textcolor{orange}{0}$ & $\textcolor{blue}{-}\textcolor{orange}{0}$ & $\textcolor{blue}{-}\textcolor{orange}{0}$ & $-3$ \\
     \cline{2-6}
     & T-LcMa       & $\textcolor{blue}{-}\textcolor{orange}{0}$ & $\textcolor{blue}{-}\textcolor{orange}{0}$ & $\textcolor{blue}{+}\textcolor{orange}{0}$ & $-1$ \\
     \cline{2-6}
     & T-FcMa       & $\textcolor{blue}{-}\textcolor{orange}{0}$ & $\textcolor{blue}{-}\textcolor{orange}{0}$ & $\textcolor{blue}{+}\textcolor{orange}{0}$ & $-1$ \\
     \cline{2-6}
     & T-PixLc      & $\textcolor{blue}{0}\textcolor{orange}{0}$ & $\textcolor{blue}{+}\textcolor{orange}{0}$ & $\textcolor{blue}{-}\textcolor{orange}{0}$ & $0$  \\
     \cline{2-6}
     & T-PixFc      & $\textcolor{blue}{+}\textcolor{orange}{0}$ & $\textcolor{blue}{+}\textcolor{orange}{0}$ & $\textcolor{blue}{-}\textcolor{orange}{0}$ & $+1$ \\
     \cline{2-6}
     & T-PixLcMa    & $\textcolor{blue}{0}\textcolor{orange}{0}$ & $\textcolor{blue}{+}\textcolor{orange}{0}$ & $\textcolor{blue}{+}\textcolor{orange}{0}$ & $+2$ \\
     \cline{2-6}
     & T-PixFcMa    & $\textcolor{blue}{+}\textcolor{orange}{0}$ & $\textcolor{blue}{+}\textcolor{orange}{0}$ & $\textcolor{blue}{+}\textcolor{orange}{0}$ & $+3$ \\
     \cline{2-6}
     & S-O          & $\textcolor{blue}{0}\textcolor{orange}{-}$ & $\textcolor{blue}{0}\textcolor{orange}{-}$ & $\textcolor{blue}{0}\textcolor{orange}{-}$ & $-3$ \\
     \cline{2-6}
     & C-Fc         & $\textcolor{blue}{-}\textcolor{orange}{-}$ & $\textcolor{blue}{-}\textcolor{orange}{-}$ & $\textcolor{blue}{-}\textcolor{orange}{-}$ & $-6$ \\
     \cline{2-6}
     & C-PixFc      & $\textcolor{blue}{+}\textcolor{orange}{-}$ & $\textcolor{blue}{+}\textcolor{orange}{-}$ & $\textcolor{blue}{-}\textcolor{orange}{-}$ & $-2$ \\
     \hline
    \end{tabularx}
    \label{tab:comparisons_SM}
\end{table}

The constraints that we implement affect the practicality of the final adversarial meshes, expressed through the production cost (PC), the difficulty of installation (DI), and difficulty of operation (DO).
Production cost refers to the estimated cost of producing the physical camouflage, including material, printing expenses, and labor time.
Difficulty of installation refers to how easy or difficult it is to physically apply or set up the camouflage on a vehicle.
Difficulty of operation assesses the extent to which the camouflage affects the normal operation or mobility of the vehicle.
See detailed discussions below.

\subsection{Texture-Based Attacks}
\label{sec:SM:practicality_and_comparisons:texture-based_attacks}

We first consider the texture modifications and their effect on the practicality.

The \textbf{Spatial Resolution} constraint (abbreviated as ``Pix'') provides a practical approach to camouflage implementation by utilizing stickers or painting squares rather than applying the entire camouflage in one go (for example, by using vinyl wraps, as discussed below). This method enhances the DI score, resulting in improved outcomes. Consequently, camouflages adhering to the ``Pix'' constraint receive a positive texture DI score ($+\text{X}$), whereas those that do not adhere to it receive a negative score ($-\text{X}$). Here, $\text{X}$ is a placeholder for the shape-related score.

Secondly, the \textbf{Spatial Restriction} constraint (referred to as ``Ma'') takes into consideration the potential challenges of operating a vehicle covered entirely by camouflage, which can restrict the vehicle's mobility. Our findings indicate that full-coverage camouflages are more effective (refer to Table 2 in the main paper; attacks involving ``Ma'' consistently yield lower EASR compared to their non-``Ma'' counterparts). However, such camouflages are only suitable for stationary vehicles, limiting operational flexibility. Introducing this constraint allows for maintaining mobility. Therefore, camouflages adhering to this constraint receive a positive texture DO score ($+\text{X}$), while those not adhering to it receive a negative score ($-\text{X}$).

The \textbf{Color Restriction} (denoted as ``Lc'' or ``Fc'') constraint minimizes the color palette for generating adversarial texture maps, impacting camouflage production costs (PC). Without any color restriction, we assume a negative texture PC score ($-\text{X}$), as full-color printing, typical for such cases, incurs high costs (e.g., starting from \$2000 for vinyl wraps).\footnote{\url{https://www.jdpower.com/cars/shopping-guides/how-much-does-it-cost-to-wrap-a-car}} Simply reducing colors does not cut costs, as vinyl wraps remain necessary. However, combining color restriction with spatial resolution (e.g., ``PixFc'') lowers costs by using stickers or manually coloring squares using a small predefined set of colors. Such combinations positively affect both PC and DI scores ($+\text{X}$). For the limited color constraint (``Lc''), where colors are automatically identified, we assume no impact on PC score ($0\text{X}$) due to potentially hard-to-obtain colors.

\subsection{Shape-Based and Combined Attacks}
The shape-based attacks do not involve any texture alterations, resulting in texture-related scores of $0$ across all three criteria. Moreover, reproducing shape modifications proves challenging, resulting in negative scores across all three shape-related criteria. Estimating the cost and difficulty of installation of such modifications remains uncertain, dependent on the vehicle's original shape and the extent of planned alterations. Similarly, assessing the difficulty of operation proves challenging and contingent on various factors. The PC, DI, and DO scores for the texture components in combined attacks mirror those of the texture-only attacks.

\subsection{Other Works}
Du \etal introduce two types of camouflages: ON and OFF. The ON type is applied on the rooftop of a vehicle, while the OFF type is placed outside of the vehicle. We find that the ON type, as well as EVD4UAV, is as practical as our most constrained texture-based attack, but its limited coverage area renders it impractical within the geospatial resolution context of our study. It could be considered as a tighter version of our implemented spatial restriction constraint (``Ma''), leading to lower performance. The OFF type of camouflage scores lower on DO due to mobility limitations. Additionally, the production cost of such camouflage remains unclear, resulting in a neutral PC texture score.

The remaining three works (FCA, ACTIVE, and DTA) share similarities with our T-Ma camouflage. Therefore, we assign them the same scores as the T-Ma camouflage.
\section{Analysis of Adversarial Textures}
\label{sec:supp_OtherPatchesAnalysis}

\begin{figure}
    \centering
    \includegraphics[width=\columnwidth]{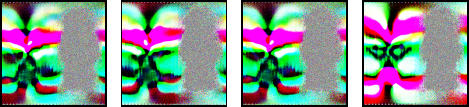}
    \caption{Examples of T-U textures obtained with different initializations. The grey region maps to the underside of the car which is ignored by the adversarial optimization.}
    \label{fig:variability_textures}
\end{figure}

Throughout our experiments, we observed a striking similarity in the prevalence of highly saturated colors, between unconstrained adversarial texture maps and adversarial patches generated in prior studies, such as those mentioned in [22,4,11]. Further analysis of our results and those of other researchers revealed that adversarial texture maps or patches generated in setups with minimal constraints tend to saturate colors located at the edges of the RGB color cube. 
They consistently exhibited extreme color saturation. \Cref{fig:variability_textures} shows the different T-U textures obtained with different attack initializations. As you can see, despite different initializations, they are very similar.

\begin{figure}
    \centering
    \includegraphics[width=\columnwidth]{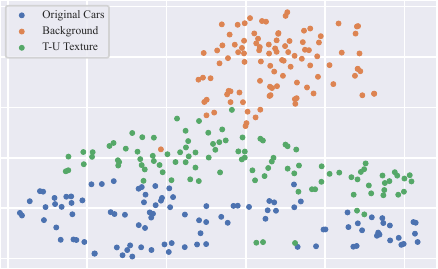}
    \caption{Embeddings of background images and vehicles with original and T-U texture maps.}
    \label{fig:embeddings}
\end{figure}

Additionally, our analysis of the latent space indicated that adversarial attacks could shift vehicle embeddings toward the background distribution but were unable to achieve complete blending, see \Cref{fig:embeddings}. We used PCA\footnote{\url{https://scikit-learn.org/stable/modules/generated/sklearn.decomposition.PCA.html}} and t-SNE\footnote{\url{https://jmlr.org/papers/v9/vandermaaten08a.html}} on features from a synthetic RetinaNet model. Replicating the background underneath the car would be the most optimal solution resulting in the perfect camouflage. As a result, the features extracted from adversarial vehicles did not closely resemble the original vehicle or the background embeddings but instead fell somewhere in between.

% \begin{figure}
%     \centering
%     \includegraphics[width=\columnwidth]{Figures/VariabilitySamples.pdf}
%     \caption{Examples of T-U textures obtained with different initializations. The grey region maps to the underside of the car which is ignored by the adversarial optimization.}
%     \label{fig:variability_textures}
% \end{figure}

% \begin{figure}
%     \centering
%     \includegraphics[width=\columnwidth]{Figures/EmbeddingFeatures.pdf}
%     \caption{Embeddings of background images and vehicles with original and T-U texture maps.}
%     \label{fig:embeddings}
% \end{figure}

As mentioned, some other works produce adversarial patches with highly saturated colors, similar to our T-U texture map. Therefore, we analyze the color distribution to verify that the colors appear at the edges of the RGB cube. To do so, we plot the distribution of pixel values along the red, green, and blue channels for each texture map.

\begin{figure*}
    \centering
    \includegraphics[width=0.8\textwidth]{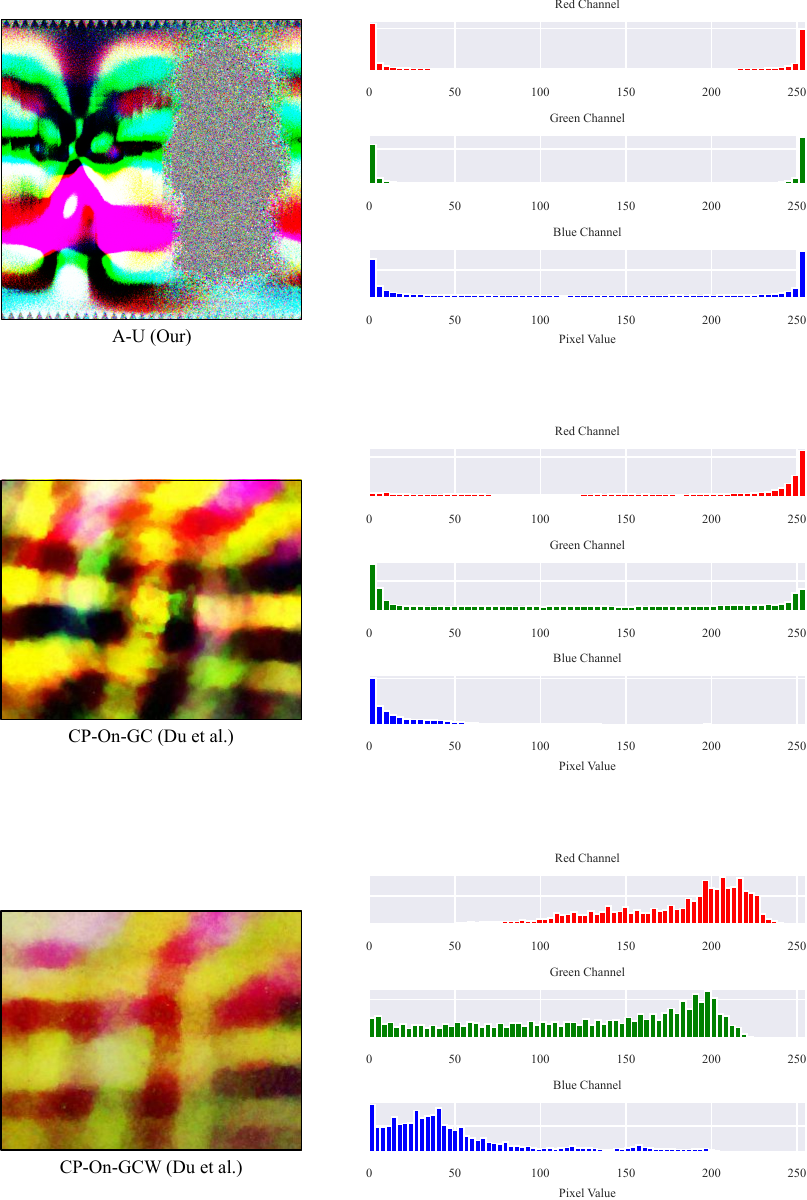}
    \caption{Color distribution in adversarial patches.}
    \label{fig:adv_patch_analysis}
\end{figure*}

Du \etal make their adversarial patches public\footnote{\url{https://github.com/andrewpatrickdu/adversarial-yolov3-cowc}} in good quality, so we use them to compare. See the results of the comparison in \Cref{fig:adv_patch_analysis}. Interestingly, they conclude that weather augmentations do not considerably improve the results. However, we find that weather augmentations during optimization significantly affect the distribution of colors by pushing a significant fraction of pixels away from the edges of the RGB cube. This type of effect can be used to constrain the optimized space, which, without any strict constraints, seems to attempt to drive the optimization outside the RGB cube (hence causing crowding at the edges).


\end{document}